\documentclass[final]{cvpr}

\usepackage{times}
\usepackage{epsfig}
\usepackage{graphicx}
\usepackage{amsmath}
\usepackage{amssymb}
\usepackage{enumitem}

\usepackage[pagebackref=true,breaklinks=true,colorlinks,bookmarks=false]{hyperref}

\usepackage{booktabs,subcaption,amsfonts,dcolumn}
\usepackage{float}
\usepackage{comment}
\usepackage{array}
\usepackage{amsmath}

\usepackage{epstopdf}
\usepackage{multirow}
\usepackage{xcolor}
\usepackage[switch]{lineno}  %
\newcolumntype{L}[1]{>{\raggedright\let\newline\\\arraybackslash\hspace{0pt}}m{#1}}
\newcolumntype{C}[1]{>{\centering\let\newline\\\arraybackslash\hspace{0pt}}m{#1}}
\newcolumntype{R}[1]{>{\raggedleft\let\newline\\\arraybackslash\hspace{0pt}}m{#1}}

\makeatletter\renewcommand\paragraph{\@startsection{paragraph}{4}{\z@}
  {.5em \@plus1ex \@minus.2ex}{-.5em}{\normalfont\normalsize\bfseries}}\makeatother

\newcolumntype{x}[1]{>{\centering\arraybackslash}p{#1pt}}
\newlength\savewidth\newcommand\shline{\noalign{\global\savewidth\arrayrulewidth\global\arrayrulewidth 1pt}\hline\noalign{\global\arrayrulewidth\savewidth}}

\def\cvprPaperID{****} 
\def\confYear{CVPR 2021}

\begin{document}

\title{Mask Guided Matting via Progressive Refinement Network}

\makeatletter
\renewcommand{\paragraph}{%
  \@startsection{paragraph}{4}%
  {\z@}{0.3ex \@plus 1ex \@minus .1ex}{-1em}%
  {\normalfont\normalsize\bfseries}%
}
\makeatother

\author{
Qihang Yu\textsuperscript{1$\ast$}~~~~
Jianming Zhang\textsuperscript{2}~~~~
He Zhang\textsuperscript{2}~~~~
Yilin Wang\textsuperscript{2}~~~~\\
Zhe Lin\textsuperscript{2}~~~~
Ning Xu\textsuperscript{2}~~~~
Yutong Bai\textsuperscript{1}~~~~
Alan Yuille\textsuperscript{1}~~~~\vspace{.3em}\\
\textsuperscript{1} The Johns Hopkins University \qquad
\textsuperscript{2} Adobe}

\twocolumn[{%
\renewcommand\twocolumn[1][]{#1}%
\maketitle
\begin{center}
 \centering
 \small
 \setlength{\tabcolsep}{0.0pt}
 \begin{tabular}{cccccc}
     \includegraphics[width=0.166\textwidth]{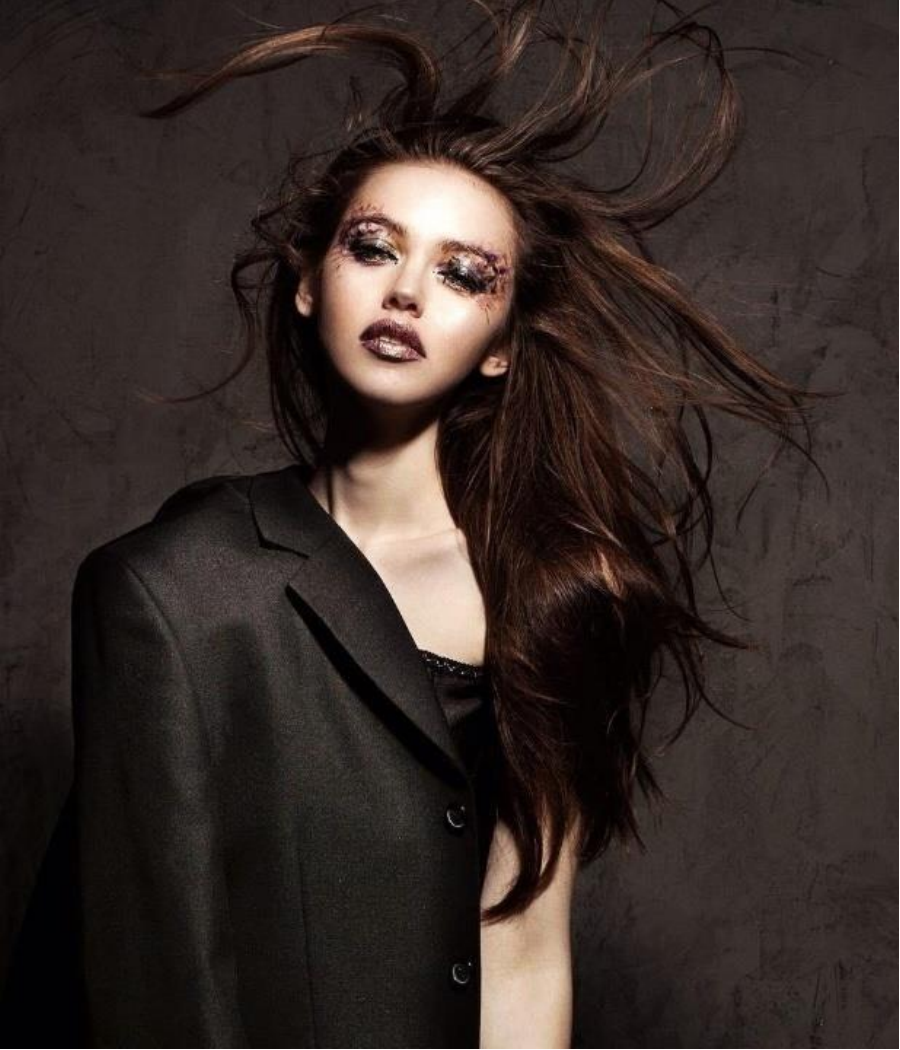} & 
     \includegraphics[width=0.166\textwidth]{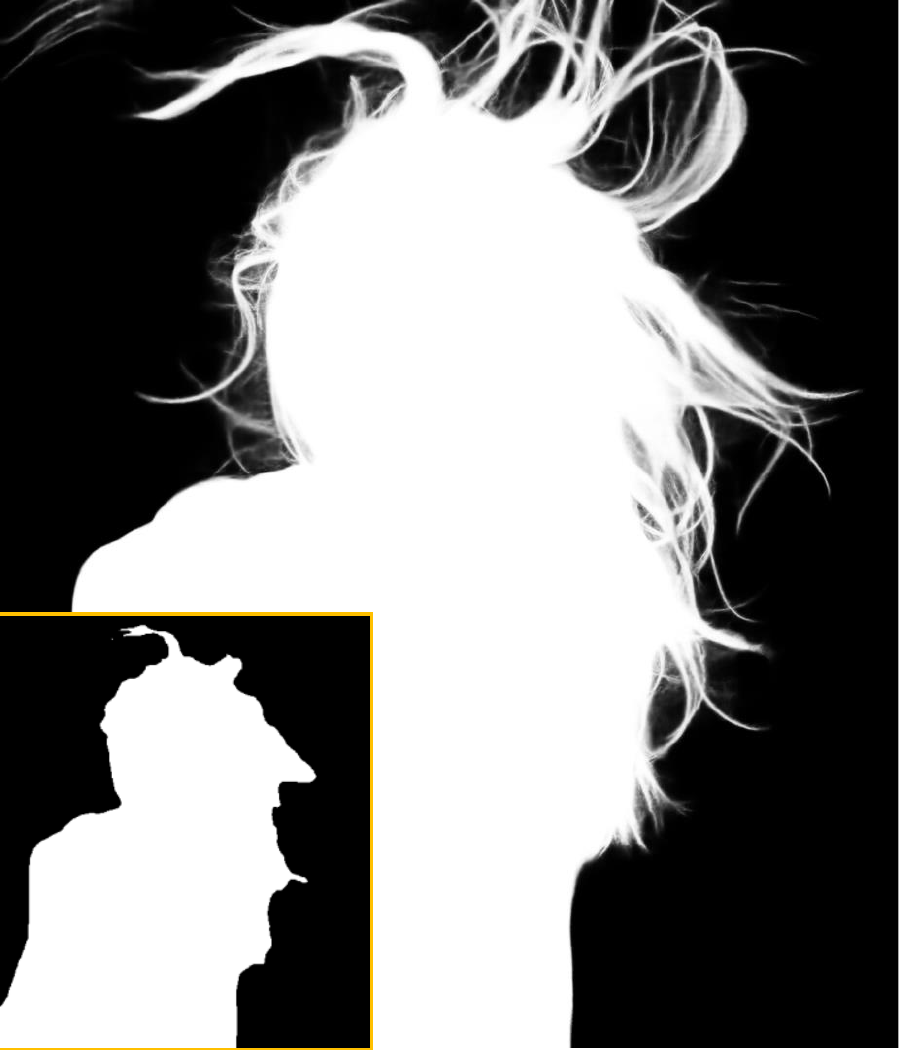} &
     \includegraphics[width=0.166\textwidth]{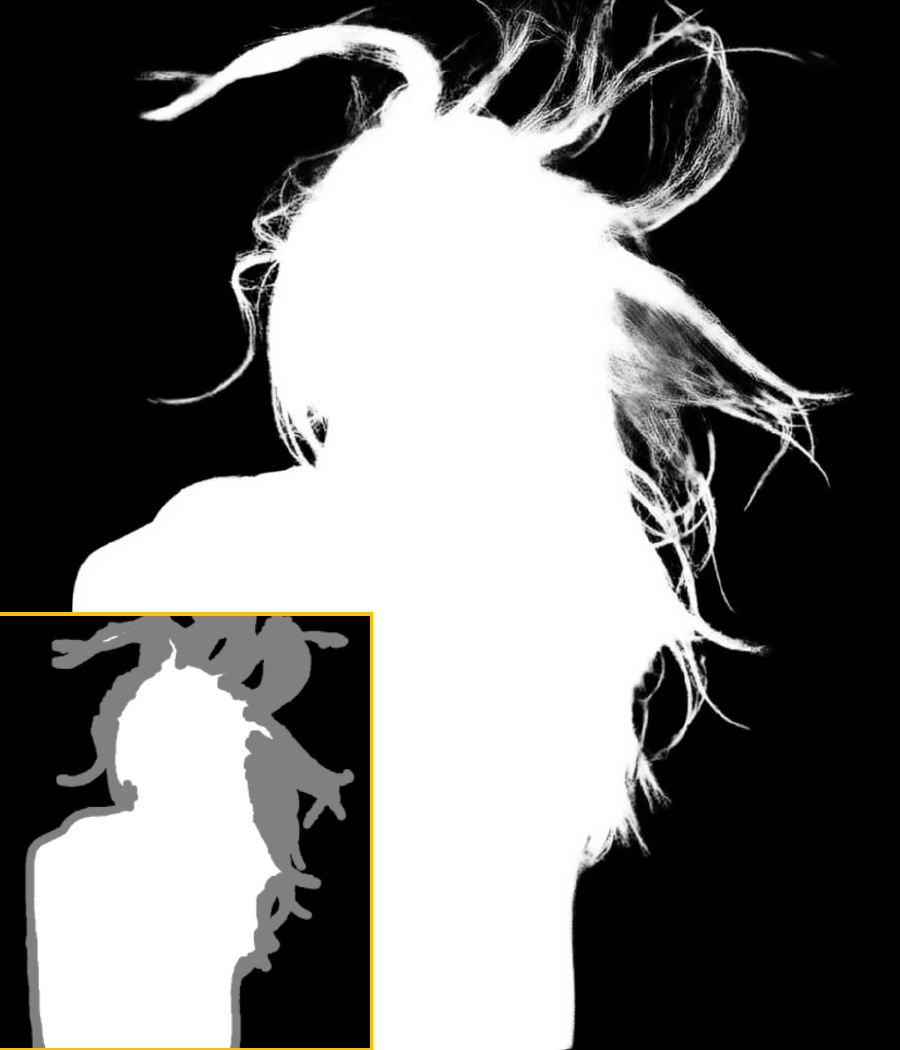} &
     \includegraphics[width=0.166\textwidth]{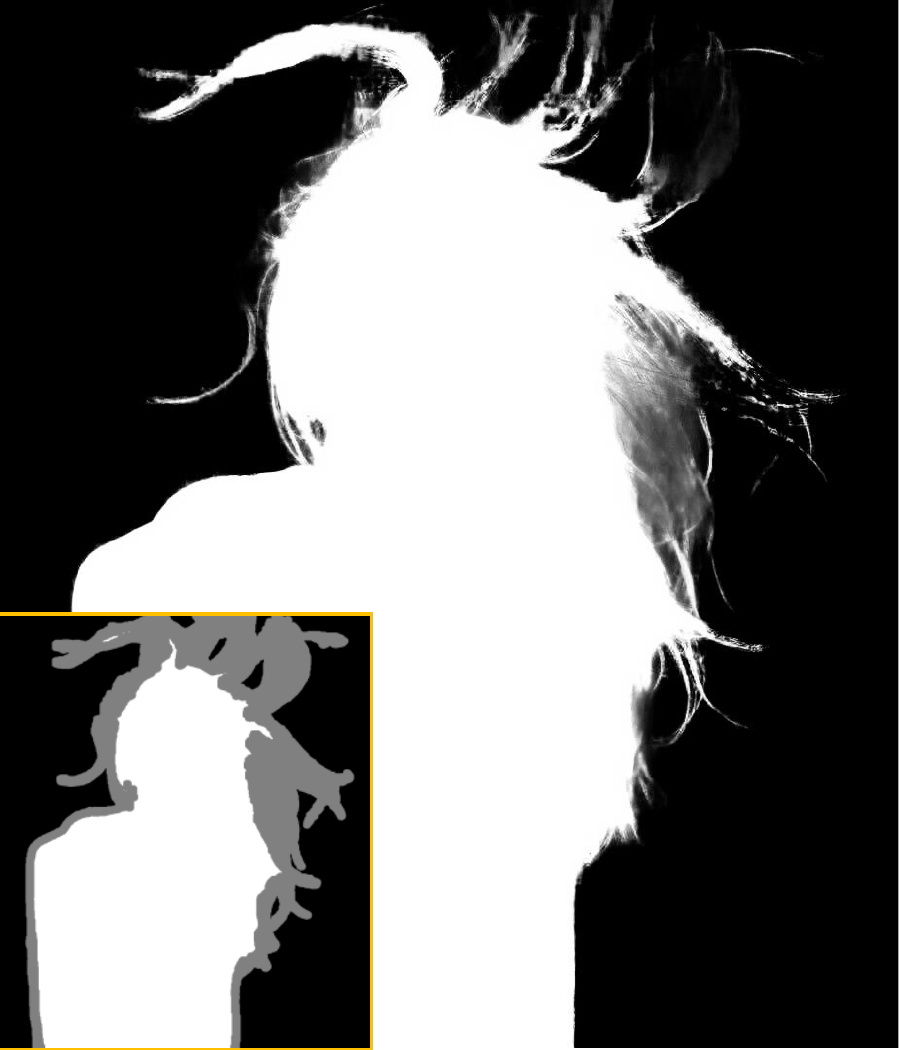} &
     \includegraphics[width=0.166\textwidth]{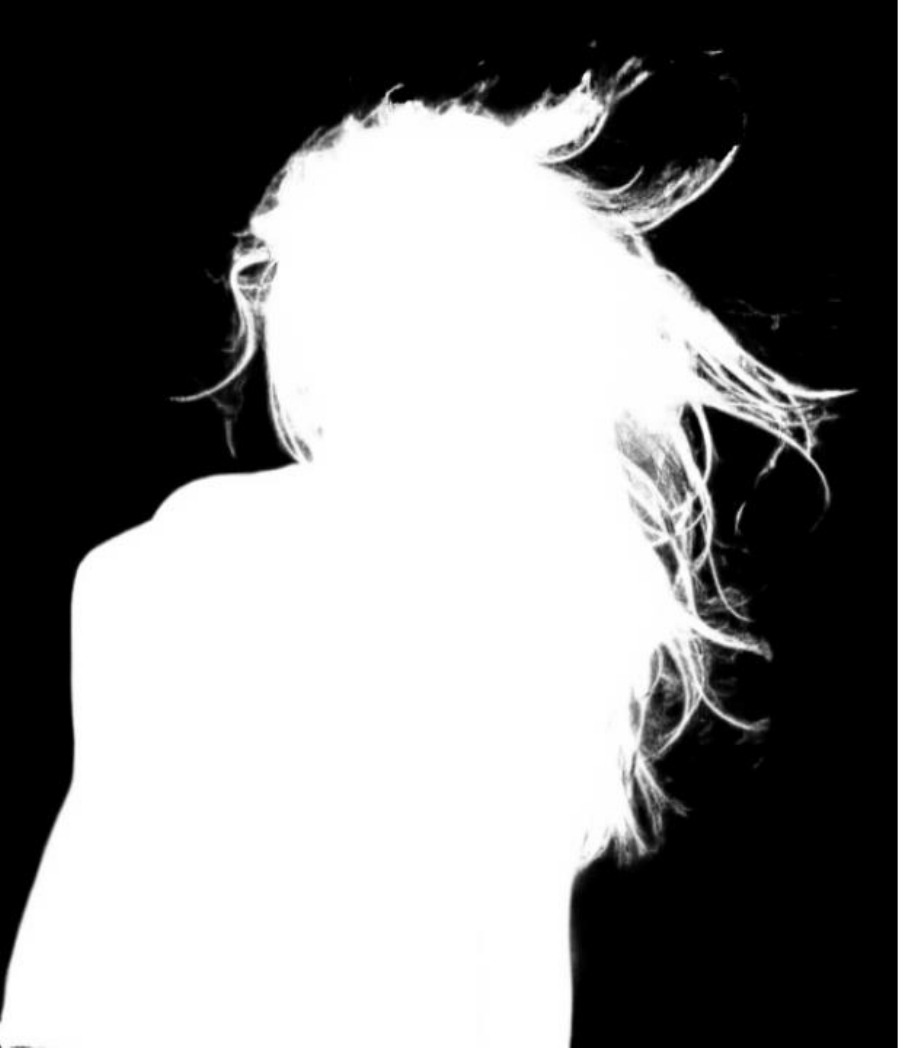} &
     \includegraphics[width=0.166\textwidth]{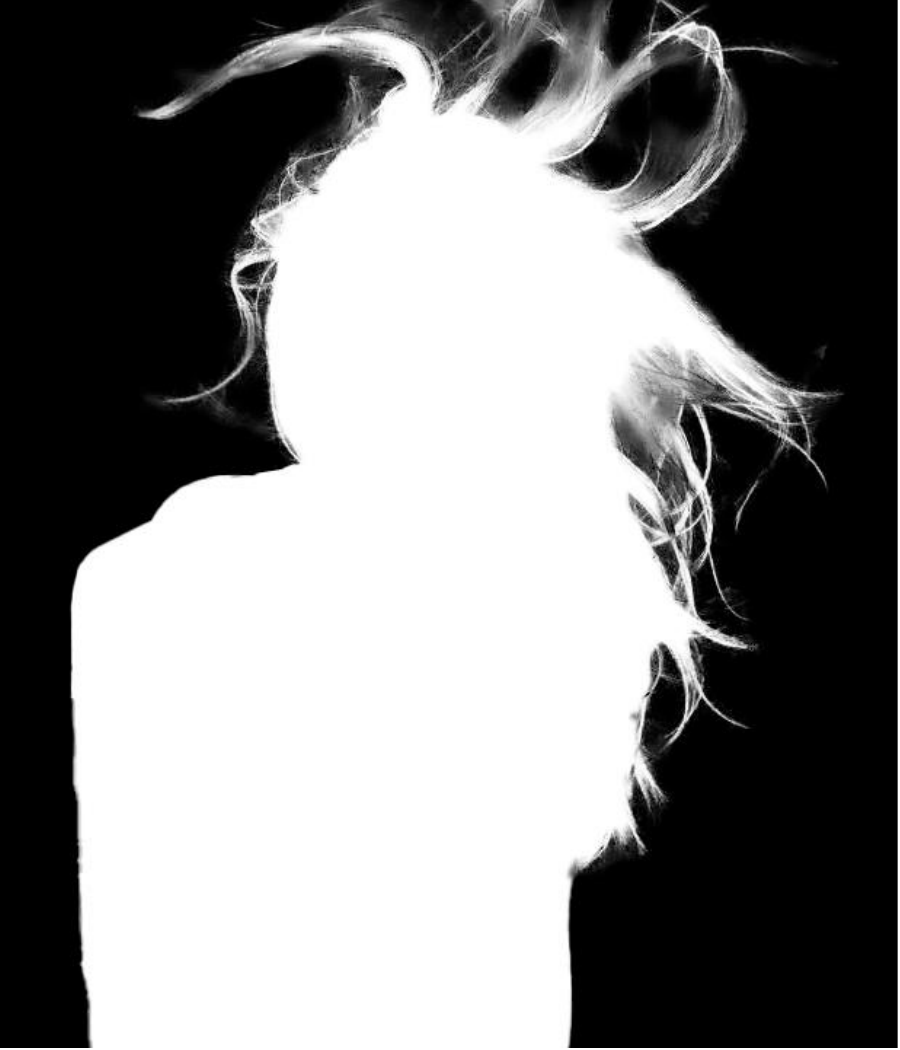} 
     \\
     Input &
     MG (ours) &
     CA \cite{hou2019context} &
     GCA \cite{li2020natural} &
     BSHM \cite{liu2020boosting} &
     Photoshop
 \end{tabular}
 \captionof{figure}{A visual comparison of MG and other matting methods including the commercial matting method in Photoshop. The guidance input (see Sec.~\ref{Experiments_portrait} for details.) is located at the bottom-left of each image. Note that BSHM~\cite{liu2020boosting} has an internal segmentation prediction network thus does not take external mask. Best viewed zoomed in.}
 \label{fig:Teaser}
\end{center}%
}]

\maketitle
\begin{abstract}
   We propose Mask Guided (MG) Matting, a robust matting framework that takes a general coarse mask as guidance.
   MG Matting leverages a network (PRN) design which encourages the matting model to provide self-guidance to progressively refine the uncertain regions through the decoding process. A series of guidance mask perturbation operations are also introduced in the training to further enhance its robustness to external guidance. We show that PRN can generalize to unseen types of guidance masks such as trimap and low-quality alpha matte, making it suitable for various application pipelines.
   In addition, we revisit the foreground color prediction problem for matting and propose a surprisingly simple improvement to address the dataset issue. 
   Evaluation on real and synthetic benchmarks shows that MG Matting achieves state-of-the-art performance using various types of guidance inputs. Code and models are available at \url{https://github.com/yucornetto/MGMatting}.\let\thefootnote\relax\footnote{$^\ast$Work done during an internship at Adobe.}
\end{abstract}
\section{Introduction}
\label{Introduction}
Image matting is a fundamental computer vision problem which aims to predict an alpha matte to precisely cut out an image region. It has many applications in image and video editing~\cite{wang2008image,xu2017deep,levin2007closed}.
Most previous matting methods require a well-annotated trimap as an auxiliary guidance input~\cite{wang2008image}, which explicitly defines the regions of foreground and background as well as the unknown part for the matting methods to solve. Although such annotation makes the problem more tractable, it can be quite burdensome for users and limits the usefulness of these methods in many non-interactive applications.

Recently, researchers start to study the matting problem in a trimap-free setting. One direction is to get rid of any external guidance, and hope that the matting model can capture both semantics and details by end-to-end training on large-scale datasets~\cite{zhang2019late,qiao2020attention}. Nevertheless, these methods are faced with the generalization challenge due to the lack of semantic guidance when tested on complex real-world images.
Another line of works investigate alternatives to the trimap guidance, easing the requirement for human input \cite{liu2020boosting,sengupta2020background,hsieh2013automatic,gupta2016automatic}. For example, \cite{hsieh2013automatic,gupta2016automatic} proposed techniques for automatic trimap generation, while \cite{sengupta2020background} takes background images instead as extra inputs. However, these methods often require a very specific type of guidance they are trained with and thus become less appealing when the guidance inputs may have varied characteristics or forms. 

In this work, we introduce a Mask Guided (MG) Matting method which takes a general coarse mask as guidance. MG Matting is very robust to the guidance input and can obtain high-quality matting results using various types of mask guidance such as a trimap, a rough binary segmentation mask or a low-quality soft alpha matte.
To achieve such robustness to guidance input, we propose a Progressive Refinement Network (PRN) module, which learns to provide self-guidance to progressively refine the uncertain matting regions through the decoding process. 
To further enhance the robustness of our method to external guidance, we also develop a series of guidance mask perturbation operations including
random binarization, random morphological operations, and also a stronger perturbation CutMask to simulate diverse guidance inputs during training.

In addition to alpha matting prediction, we also revisit the foreground color prediction problem for matting. Without accurately recovering the foreground color in the transparent region, the composited image will suffer from the fringing issue. We note that the foreground color labels in the widely-used dataset~\cite{xu2017deep} are suboptimal for model training due to the labeling noise and limited diversity. As a simple yet effective solution, we propose Random Alpha Blending (RAB) to generate synthetic training data from random alpha mattes and images. We show that such simple method can improve the foreground color prediction accuracy without requiring additional manual annotations. As a result, combining with the proposed PRN, MG Matting is able to generate more visual plausible composition results.  

Our contributions can be summarized as follows:
\begin{itemize}[noitemsep]
    \item We propose Mask Guided Matting, a general matting framework working with guidance masks in various qualities and even forms, and achieve a new state-of-the-art performance evaluated on both synthetic and real-world datasets.
    \item We introduce Progressive Refinement Network (PRN) along with a guidance perturbation training pipeline as a solution to learning a robust matting model.
    \item We study the problem of foreground color prediction for matting and propose a simple improvement using random alpha blending. 
\end{itemize}
In addition, we collect and release a high-quality matting benchmark dataset of real images to evaluate the real-world performance of matting models.
\section{Related Work}
\label{RelatedWork}
\paragraph{Trimap-based Image Matting.} A majority of matting methods requires a trimap as additional input, which divides an image into foreground, background, and unknown regions. Traditional methods are often sampling-based or propagation-based. Sampling-based ones~\cite{gastal2010shared,chuang2001bayesian,he2011global,shahrian2013improving,wang2007optimized} estimate foreground/background color statistics through sampling pixels in the definite foreground/background regions to solve the alpha matte in the unknown region. The propagation-based methods~\cite{chen2013knn,lee2011nonlocal,levin2007closed,levin2008spectral,sun2004poisson,he2010fast}, also known as affinity-based methods, estimate alpha mattes by propagating the alpha value from foreground and background pixels to the unknown area.

Recently, deep learning approaches have been proved successful in many areas, including classification~\cite{he2016deep, tan2019efficientnet,li2020shape, li2020neural}, detection~\cite{he2017mask,bai2019semantic,bai2020coke}, and segmentation~\cite{chen2017deeplab,yu2020c2fnas}. It also have achieved great success in image matting.~\cite{xu2017deep} created a matting dataset with annotated mattes composited to various background images, and trained a deep network on it. Later,~\cite{lutz2018alphagan} introduced a generative adversarial framework to improve the results. \cite{tang2019learning} proposed to combine the sampling-based method and deep learning.~\cite{lu2019indices} introduced a new index-guided upsampling and unpooling operations to better keep details in the predictions. \cite{hou2019context} proposed a two-encoder two-decoder architectures to simultaneous estimate foreground and alpha. \cite{li2020natural} further boost the performance with a contextual attention module. 

\paragraph{Trimap-free Image Matting.} It is noticeable that there are also some trials~\cite{aksoy2018semantic,shen2016deep} to get rid of the trimap to predict alpha matte. \cite{zhang2019late} proposed a framework consisting of a segmentation network and a fusion network, where the input is only a single RGB image. Later, \cite{liu2020boosting} introduced a trimap-free framework consisting of mask prediction network, quality unification network, and matting refinement network for human portrait matting. The trimap-free matting performance is further boosted with attention module~\cite{qiao2020attention}. However, these trimap-free methods still have some gap to trimap-based ones in terms of performance. Another direction is to use an alternative guidance to trimap. \cite{sengupta2020background} introduced a framework taking background images along with other potential priors (\emph{e.g.}, segmentation mask, motion cue) as additional inputs. It shows great potential and can obtain a comparable performance to state-of-the-art trimap-based methods.

\begin{figure*}[!t]
    \centering
    \includegraphics[width=0.9\linewidth]{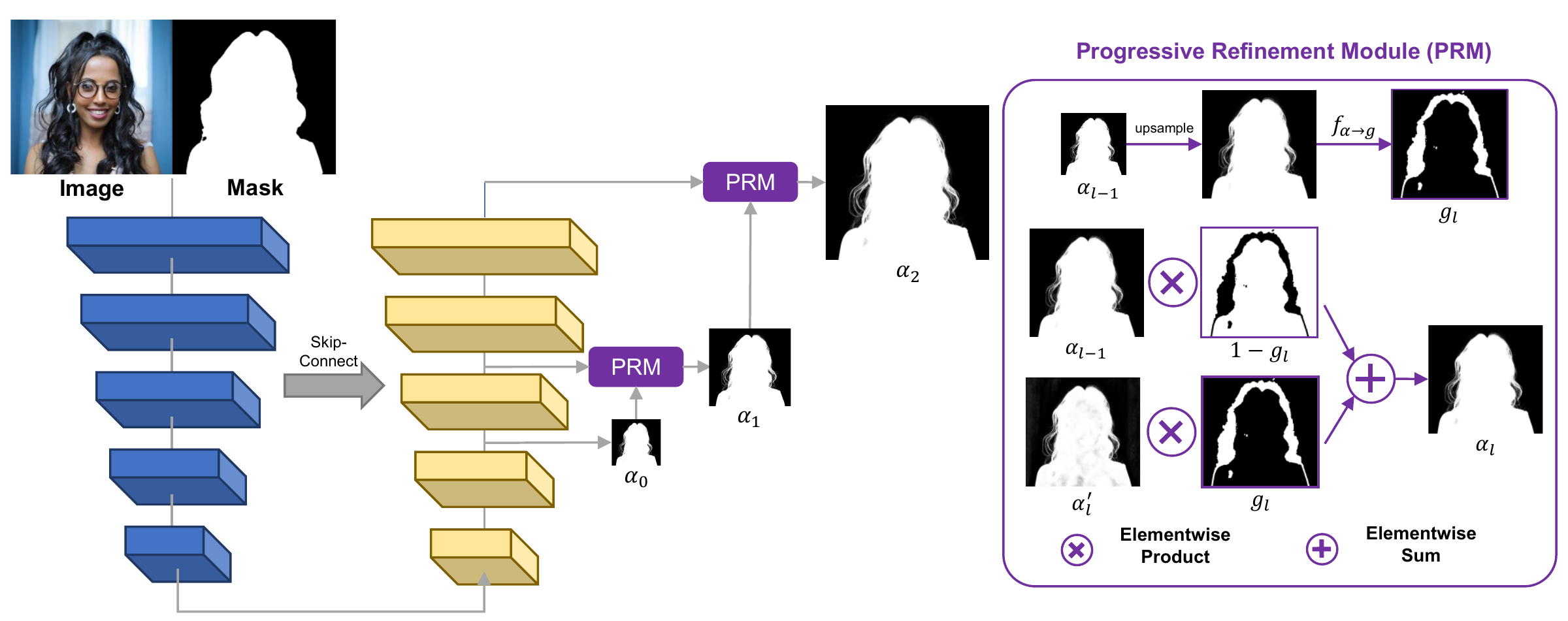}
    \caption{The proposed PRN. The network predicts alpha matte at multiple resolutions, while the one at lower-resolution provides guidance about uncertain region to be refined in the next prediction.} 
    \label{fig:Architecture}
\end{figure*}

\paragraph{Foreground Color Decontamination.} Many conventional matting methods~\cite{gastal2010shared,levin2007closed} proposed to predict both alpha matte and foreground color for extracting foreground objects. However, it is only very recently~\cite{hou2019context} incorporated the foreground prediction into the deep learning framework. Later,~\cite{sengupta2020background} also predicts foreground color to reduces artifacts for a better composition result. Nevertheless, these methods mainly add a foreground decoder and directly learn from color label in~\cite{xu2017deep}, which only provides limited training samples and, more seriously, the color labels can be inaccurate and noisy(see Fig.~\ref{fig:color_bias}).~\cite{forte2020f} proposes to use~\cite{levin2007closed} to obtain a smoother color label. 

Our method differs from algorithms mentioned above in the following folds: 1) Our model works in a more general setting where only an easy-to-obtain coarse mask, no matter user-defined or model-predicted, is needed as guidance. It could handle different qualities and even various types of guidance as input. Thus it could be used as either trimap-based or trimap-free model depending on what guidance is available. Our model could also leverage a stronger guidance to achieve even finer details. 2) Our methods could also predict the foreground color. Unlike~\cite{hou2019context}, where the foreground prediction is directly learned from the color label, we note that the limited training data and inaccurate human label result in undesired results especially in the boundary regions.
Instead, we propose to use Random Alpha Blending to avoid the bias in label, which not only introduces more diverse training samples but also avoid the inaccurate color label locating in boundary regions.

\section{MG Matting}
\label{Method}

The problem of image matting can be formulated as:
\begin{equation}
\label{eqn:matting}
    \mathbf{I} = \mathbf{\alpha}\mathbf{F} + (1-\mathbf{\alpha})\mathbf{B}, \mathbf{\alpha}\in [0,1],
\end{equation}
where $\mathbf{I}$, $\mathbf{F}$, $\mathbf{B}$, and $\mathbf{\alpha}$ refer to the image color, foreground color, background color and alpha matte respectively. As only $\mathbf{I}$ is observed, this is a very ill-posed problem. To solve the matting problem, most methods require a trimap input, which labels the foreground region (\ie $\mathbf{\alpha} = 1$), the background region (\ie $\mathbf{\alpha} = 0$) and the unknown part. In practice, the trimap input can contain various levels of noise and errors, making the matting results inconsistent.

We relax the strong assumption of the trimap by proposing a Mask Guided Matting method. The mask guidance, such as a predicted segmentation mask or a rough manual selection, only provides a coarse spatial prior of the foreground region. Therefore, our MG Matting method needs more high-level semantic understanding of the input mask, so that it can detect the foreground/background region and the soft transparent part robustly. Meanwhile, our model has to capture image low-level patterns such as edge and texture to produce fine details of the target matte. Coordinating the high-level and the low-level feature learning is the key to the design of our MG Matting method.

To this end, we introduce Progressive Refinement Network (PRN), which provides a coarse-to-fine self-guidance to progressively refine the uncertain regions during the decoding process. 
In the following, we present the details of PRN, the training formulation and some data augmentation techniques to enhance the robustness of our model.

\subsection{Progressive Refinement Network}

An overview of the PRN is shown in Fig.~\ref{fig:Architecture}. The structure of our PRN follows the popular encoder-decoder network with skip connections. Our network takes an image and a coarse mask as input and outputs a matte. During the decoding process, PRN has a side matting output at each feature level. The side outputs with deep supervision have been shown to improve the feature learning at different scales~\cite{xie2015holistically}. However, unlike~\cite{xie2015holistically}, we find that linearly fusing the side outputs is not ideal for the matting problem (see Table~\ref{tab:DIM_Ablation} for details). This is because image region closer to the object boundary requires lower-level features to delineate the foreground, while identifying internal object regions needs higher-level guidance.

To address this problem, we introduce a Progressive Refinement Module (PRM) at each feature level to selectively fuse the matting outputs from the previous level and the current level. Specifically, for the current level $l$ we generate a self-guidance mask $g_{l}$ from the matting output $\alpha_{l-1}$ of the previous level using the following function:
\begin{equation}
\label{eqn:alpha2guidance}
f_{\alpha_{l-1}\rightarrow g_l}(x,y)=\left\{
\begin{array}{rcl}
1     &      & \mathrm{if~} {0 < \alpha_{l-1}(x,y) < 1},\\
0    &      & \mathrm{otherwise}.
\end{array} \right. 
\end{equation}
The $\alpha_{l-1}$ is firstly upsampled to match the size of the raw matting output $\alpha_{l}'$ of the current level and then produces resultant self-guidance mask $g_l$.
The self-guidance mask defines the transparent region (\ie $0<\alpha<1$) as unknown and replaces the unknown region of $\alpha_{l-1}$ with the current raw output $\alpha_{l}'$ to obtain an updated $\alpha_{l}$ of current level:
\begin{equation}
\label{eqn:selfrefine}
    \alpha_{l} = \alpha_{l}'g_{l}+\alpha_{l-1}(1-g_{l}).
\end{equation}
In this way, confident regions predicted from the previous higher-level features are preserved and the current level only needs to focus on refining the uncertain region. 

In practise, we obtain alpha matte side outputs at three feature levels of stride 8, 4, and 1 respectively (see Fig.~\ref{fig:Architecture}) and slightly dilate the self-guidance masks for a more robust self-guidance.
The initial base matte of $1/8$ image size will be progressively upsampled and refined, and the uncertain regions will also shrink gradually through the decoding process using the proposed PRM. The full network is trained end-to-end to auto-balance the refinement focus at multiple feature levels. Such self-guided refinement also makes model less reliant on the external mask guidance, leading to more robust matting performance.

\paragraph{Training scheme.} 
For loss functions, we adopt the $l_1$ regression loss, composition loss~\cite{xu2017deep}, Laplacian loss~\cite{hou2019context} and denote them as $\mathcal{L}_{l1}$, $\mathcal{L}_{comp}$, $\mathcal{L}_{lap}$ respectively. We represent the ground truth alpha with $\hat{\alpha}$ and prediction alpha with $\alpha$. The overall loss functions is the summation of them:
\begin{equation}
    \mathcal{L}(\hat{\alpha}, \alpha) =  \mathcal{L}_{l1}(\hat{\alpha}, \alpha) + \mathcal{L}_{comp}(\hat{\alpha}, \alpha) + \mathcal{L}_{lap}(\hat{\alpha}, \alpha).
    \label{loss_type}
\end{equation}
The loss is applied to each output head of the network.
To make the training more focused on the unknown region, We further modulate the loss with $g_l$. The final loss function can be formulated as:
\begin{equation}
    \mathcal{L}_{final} = \sum_{l}w_l\mathcal{L}(\hat{\alpha_{l}}\cdot g_{l}, \alpha_{l} \cdot g_{l}),
    \label{loss_final}
\end{equation}
where $w_l$ is the loss weight assigning to the outputs of different levels. We use $w_0:w_1:w_2=1:2:3$ in our experiments. $g_{l}$ is generated from $\alpha_{l-1}$ by Eqn.~\ref{eqn:alpha2guidance}, and $g_{0}$ is a mask filled with one so that the base level output can be supervised over the whole image to provide  more holistic semantic guidance for the next level output. 

For data augmentation, we follow the training protocol proposed in~\cite{li2020natural}, including random composite two foreground object images, random resize images with random interpolation methods, random affine transformation, color jitters. We random crop $512\times512$ patches centered on an unknown region for training. Each patch is composited to a random background image from MS COCO dataset~\cite{lin2014microsoft}.

\paragraph{Guidance Perturbation.} To ensure that our model can adapt to guidance masks from different sources and with different qualities, we propose a series of guidance perturbation to generate guidance masks from ground-truth alpha matte during training. Given a ground-truth alpha matte, we first binarize it with a random threshold uniformly sampled from $0$ to $1$. Then, the mask is dilated and/or eroded in random order with random kernel sizes from $1$ to $30$.

Moreover, we provide a stronger guidance perturbation named CutMask to further improve the model robustness. Inspired by the successful natural image augmentation CutMix~\cite{yun2019cutmix}, we randomly select a patch size ranging from $1/4$ to $1/2$ image size. Then, two random patches of the guidance are selected and the content of one patch will overwrite another. This stronger perturbation provides additional localized guidance mask corruption, making the model more robust to semantic noises in external guidance masks.

Besides perturbing external guidance masks, we note that perturbing internal self-guidance mask is also very important to improve the robustness. Therefore, we randomly dilate the self-guidance masks to incorporate more variance. Particularly, during training, the self-guidance mask from output stride 8 is dilated by $K_1$ random sampled from $[1,30]$ and the one from output stride 4 is dilated by $K_2$ from $[1,15]$. For testing, we fix $K_1=15$ and $K_2=7$.

\begin{figure}[!t]
    \centering
    \includegraphics[width=1.0\linewidth]{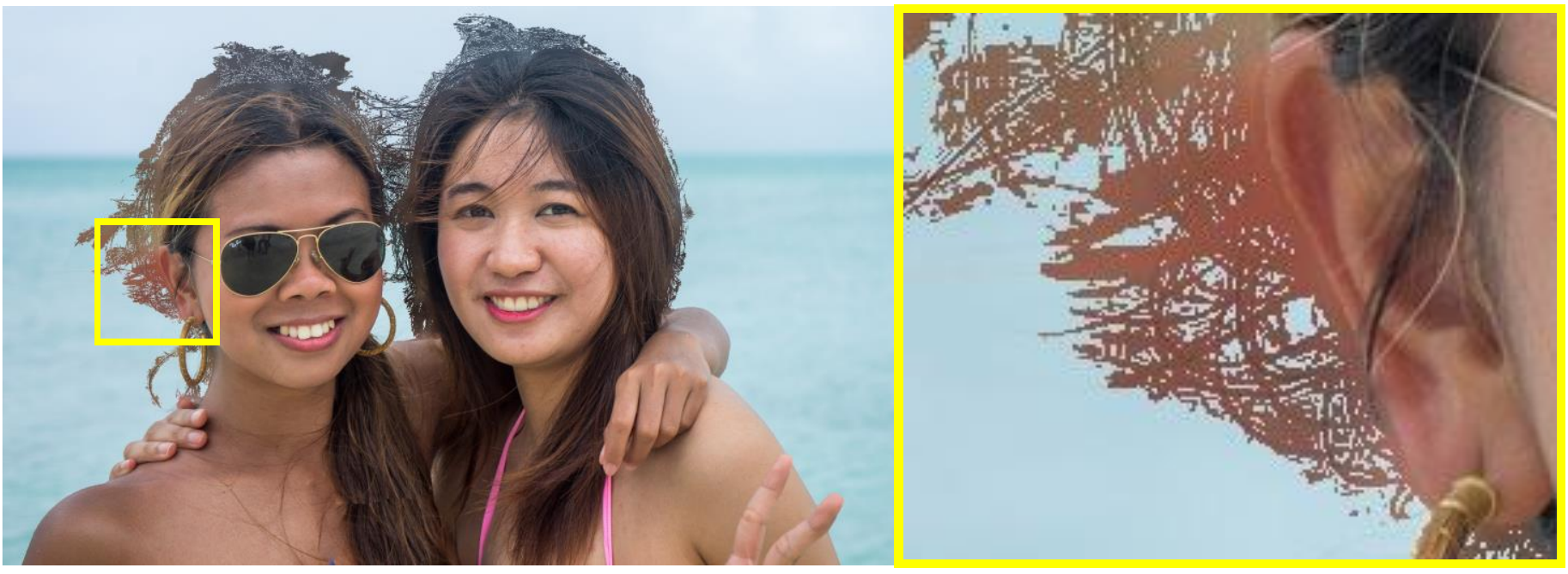}
    \caption{The color labels in the commonly used training data from ~\cite{xu2017deep} are noisy and inaccurate especially near the boundary part. Note that the hair near the ear falsely gets pinker. Best viewed in color and zoomed in.}
    \label{fig:color_bias}
\end{figure}

\subsection{Foreground Color Estimation}
As indicated in Eqn.~\ref{eqn:matting}, both alpha matte and foreground color need to be solved for foreground object extraction. Nevertheless, only a few matting methods learn to predict the foreground color~\cite{hou2019context,sengupta2020background} and all of them used the popular Composition-1k dataset~\cite{xu2017deep} for training. 

However, there are a couple of issues in the Composition-1k dataset. First of all, this dataset only contains 431 foreground images with matting and foreground color ground truth, which is quite limited to train a foreground color model. Moreover, the foreground color labels, which were estimated using the color decontamination feature in Photoshop~\cite{xu2017deep}, are sometimes noisy and inaccurate near the boundary regions (see Fig.~\ref{fig:color_bias}). This can introduce color spills and other artifacts into the images during data augmentation process, making the learning less stable. Besides, labels are only provided where the alpha value is greater than zero, so existing methods can only apply supervision to the foreground region~\cite{hou2019context}, leading to unstable behaviors in the undefined part.

To address these issues, we propose a simple yet effective method, named Random Alpha Blending (RAB), to generate synthetic training data by blending a foreground image and a background image using a randomly selected alpha matte.
Although the composited images may not be semantically meaningful, they can provide accurate and unbiased foreground color labels in the transparent region. The random alpha blending can also significantly make training data more diverse and improve the generalization of the foreground color prediction. Besides, we also note that RAB makes it possible to apply loss supervision over all image, leading to a much smoother prediction which is desired for robust compositing. (See Fig.~\ref{fig:color_compare})

For foreground estimation, we train a separate model using a basic encoder-decoder network, which takes an image and an alpha matte as input. The loss function is the summation of $l_1$ regression loss, compositing loss, and Laplacian loss. We note that although training a single model for both matte and foreground color prediction is possible, empirically this will degrade the matting performance~\cite{hou2019context}, and the random alpha blending will destroy the semantic cue for the matting model. In addition, decoupling foreground color prediction from matting makes the color model transferable to the use cases where the matte is already given.
\begin{table}[tb]
\setlength{\tabcolsep}{2.0pt}
\centering
\footnotesize
\begin{tabular}{L{0.46\linewidth}|C{0.11\linewidth}C{0.11\linewidth}C{0.11\linewidth}C{0.11\linewidth}}
\shline
Methods                   & SAD                   & MSE ($10^{-3}$)             & Grad                  & Conn                  \\ \shline
Learning Based Matting~\cite{zheng2009learning}    & 113.9                 & 48                    & 91.6                  & 122.2                 \\
Closed-Form Matting~\cite{levin2007closed}       & 168.1                 & 91                    & 126.9                 & 167.9                 \\ 
KNN Matting~\cite{chen2013knn}               & 175.4                 & 103                   & 124.1                 & 176.4                 \\ 
Deep Image Matting~\cite{xu2017deep}        & 50.4                  & 14                    & 31.0                  & 50.8                  \\ 
IndexNet Matting~\cite{lu2019indices}          & 45.8                  & 13                    & 25.9                  & 43.7                  \\ 
AdaMatting~\cite{cai2019disentangled}                & 41.7                  & 10.2                  & 16.9                  & -                     \\ 
Context-Aware Matting~\cite{hou2019context}     & 35.8                  & 8.2                   & 17.3                  & 33.2                  \\ 
GCA Matting~\cite{li2020natural}               & 35.3                  & 9.1                   & 16.9                  & 32.5                  \\ \shline
Ours$_\mathrm{TrimapFG}$      & \textbf{31.5} & \textbf{6.8} & \textbf{13.5} & \textbf{27.3} \\
Ours$_\mathrm{Trimap}$             & 32.1 & 7.0 & 14.0 & 27.9 \\ \shline
\end{tabular}
\caption{Results on Composition-1k test set. The subscripts denote the corresponding guidance inputs, $\ie$ TrimapFG, Trimap. The other evaluated methods all require a trimap as input.}
\label{tab:DIM}
\end{table}

\begin{table}[tb]
\setlength{\tabcolsep}{2.0pt}
\centering
\footnotesize
\begin{tabular}{L{0.47\linewidth}|C{0.11\linewidth}C{0.11\linewidth}C{0.11\linewidth}C{0.11\linewidth}}
\shline
Methods                   & SAD                   & MSE ($10^{-3}$)             & Grad                  & Conn                  \\ \shline
Learning Based Matting$^{*}$~\cite{zheng2009learning}    & 105.04                 & 21                    & 94.16                  & 110.41                 \\
Closed-Form Matting$^{*}$~\cite{levin2007closed}       & 105.73                 & 23                    & 91.76                 & 114.55                 \\
KNN Matting$^{*}$~\cite{chen2013knn}               & 116.68                 & 25                   & 103.15                 & 121.45                 \\
Deep Image Matting$^{*}$~\cite{xu2017deep}        & 47.56                  & 9                    & 43.29                  & 55.90   
\\
HAttMatting$^{*}$~\cite{qiao2020attention}        & 48.98                  & 9                    & 41.57                  & 49.93  
\\ \shline
Deep Image Matting~\cite{xu2017deep}        & 48.73                  & 11.2                    & 42.60                  & 49.55   
\\
+ Ours        & \textbf{36.58}                  & \textbf{7.2}                    & \textbf{27.37}                  & \textbf{35.08}   
\\ \hline
IndexNet Matting~\cite{lu2019indices}          & 46.95                  & 9.4                    & 40.56                  & 46.80                  \\
+ Ours        & \textbf{35.82}                  & \textbf{5.8}                    & \textbf{25.75}                  & \textbf{34.23}   
\\ \hline
Context-Aware Matting~\cite{hou2019context}     & 36.32                  & 7.1                   & 29.49                  & 35.43                 \\ 
+ Ours        & \textbf{35.04}                  & \textbf{5.4}                    & \textbf{24.55}                  & \textbf{33.35}   
\\ \hline
GCA Matting~\cite{li2020natural}               & 39.64                  & 8.2                   & 32.16                  & 38.77                  \\ 
+ Ours        & \textbf{35.93}                  & \textbf{5.7}                    & \textbf{25.94}                  & \textbf{34.35}   
\\ \shline
\end{tabular}
\caption{Matting refinement results on Distinction-646 test set. Results with $*$ are from methods trained on Distinction-646 train set as reported in~\cite{qiao2020attention} for reference. Other results are only trained on composition-1k.}
\label{tab:Distinction}
\vspace{-1em}
\end{table}

\section{Experiments on Synthetic Datasets}
\label{Experiments_DIM}
In this section, we report the evaluation results of our method under the traditional synthetic data setting, where the test images are generated using foreground images with ground truth mattes and random background images. 

\paragraph{Evaluation Metrics.} We follow previous methods to evaluate the results by Sum of Absolute Differences (SAD), Mean Squared Error (MSE), Gradient (Grad) and Connectivity (Conn) errors using the official evaluation code~\cite{xu2017deep}. 

\label{implementation_DIM}
\paragraph{Network Architectures.} We adopt ResNet34-UNet proposed in~\cite{li2020natural} with an Atrous Spatial Pyramid Pooling (ASPP)~\cite{chen2017deeplab} as the backbone for both PRN and color prediction. The first convolution layer is adjusted to take a 4-channel input consisting of a RGB image along with an external guidance input. Moreover, an alpha prediction head (Conv-BN-ReLU-Conv) is attached to the features at output stride 4 and 8 respectively to obtain side outputs. 

\begin{figure}[!t]
    \centering
    \includegraphics[width=1.0\linewidth]{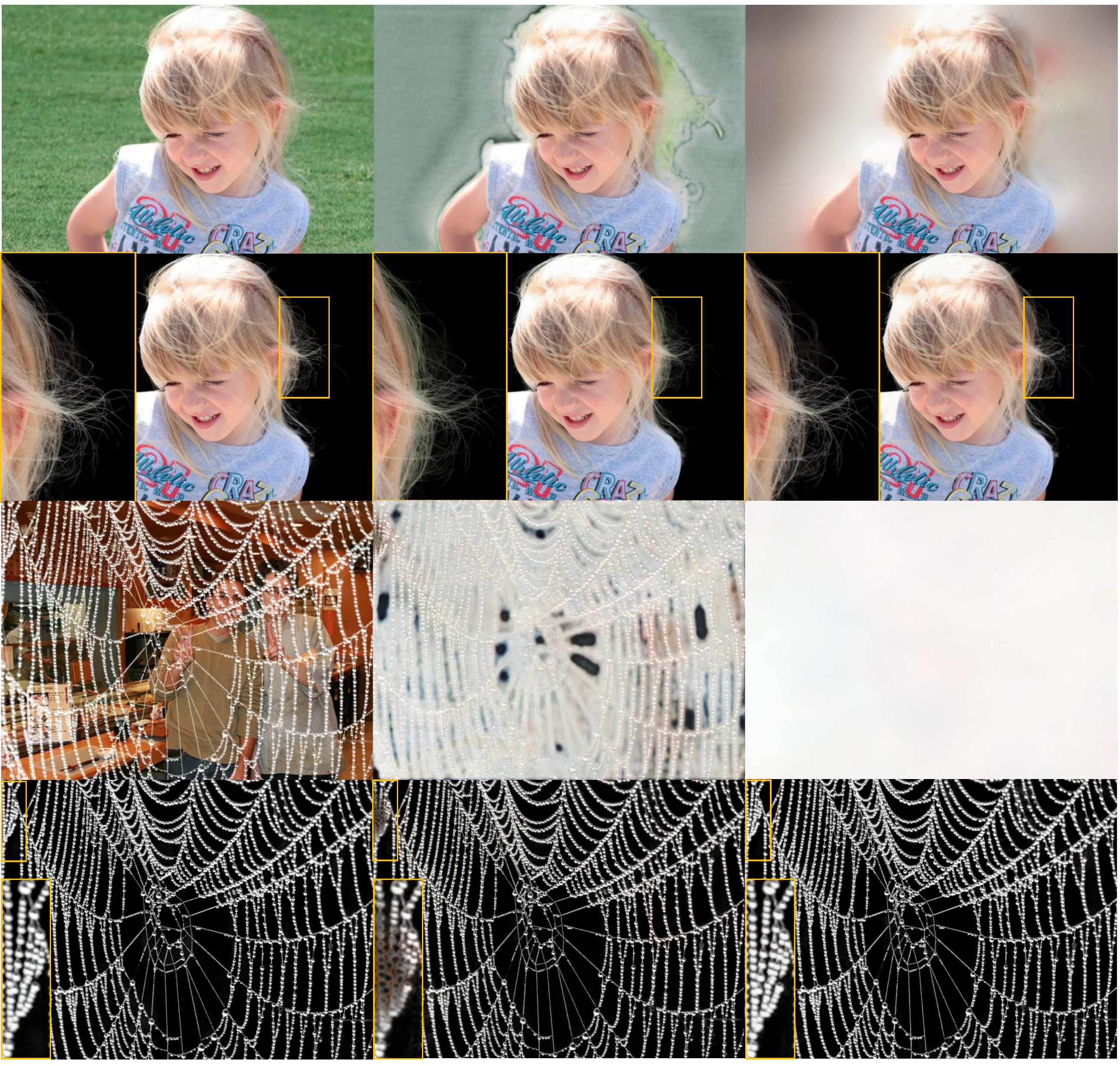}
    \caption{A visual comparison of foreground color decontamination. Each column from left to right: Input image and ground truth $\alpha\cdot F$, Foreground color prediction and $\alpha\cdot F$ of~\cite{hou2019context}, predictions of our model with random alpha blending. Note that the background color is mixed into the prediction of~\cite{hou2019context}, while our model can estimate a more smooth foreground color map and be more robust.}
    \label{fig:color_compare}
    \vspace{-1em}
\end{figure}

\paragraph{Training stage.} 
To fairly compare with previous deep image matting methods, we train our MG Matting model using the Composition-1k datasest~\cite{xu2017deep} which contains 431 foreground objects and the corresponding ground-truth alpha mattes for training.
The network is initialized with ImageNet~\cite{deng2009imagenet} pre-trained weight. We use crop size 512, batch size of 40 in total on 4 GPUs, Adam optimizer with $\beta_{1}=0.5$ and $\beta_{2}=0.999$. The learning rate is initialized to $1\times 10^{-3}$. The training lasts for $100,000$ iterations with warm-up at the first $5,000$ iterations and cosine learning rate decay~\cite{loshchilov2016sgdr,goyal2017accurate}. We also apply a curriculum learning manner to help the PRN training. Particularly, for the first $5,000$ iterations, the predictions of output stride 4 and 1 will be guided by guidance mask generated from ground-truth alpha, and for the next $10,000$ iterations, the guidance will be evenly and randomly generated from self-prediction and ground-truth alpha. Afterwards, each alpha prediction should fully rely on its self-guidance. The foreground color prediction is trained under the exactly same settings except that the generated training samples are composited by random foreground and alpha matte. It is noticeable that with RAB, we can add foreground color supervision on the whole image instead of only foreground regions, which produces more smooth and stable results (see Fig.~\ref{fig:color_compare}).

\begin{table}[tb]
\setlength{\tabcolsep}{3.0pt}
\centering
\small
\begin{tabular}{L{0.6\linewidth}|C{0.11\linewidth}C{0.11\linewidth}}
\shline
Methods               & SAD    & MSE ($10^{-3}$) \\ \shline
Global Matting~\cite{he2011global}        & 220.39 & 36.29     \\ 
Closed-Form Matting~\cite{levin2007closed}   & 254.15 & 40.89     \\ 
KNN Matting~\cite{chen2013knn}           & 281.92 & 36.29     \\ 
Context-Aware Matting~\cite{hou2019context} & 61.72  & 3.24      \\ \shline
Ours             & \textbf{49.80}      & \textbf{2.48}         \\ \shline
\end{tabular}
\caption{The foreground result ($\alpha\cdot F$) on the Composition-1k dataset.}
\label{tab:DIM_Color}

\end{table}

\begin{table}[tb]
\setlength{\tabcolsep}{1.0pt}
\centering
\small
\begin{tabular}{L{0.5\linewidth}|C{0.11\linewidth}C{0.11\linewidth}|C{0.11\linewidth}C{0.11\linewidth}}
\shline
\multicolumn{1}{c|}{\multirow{3}{*}{Methods}} & \multicolumn{2}{c|}{Whole Image} & \multicolumn{2}{c}{Unknown Area} \\ \cline{2-5} 
\multicolumn{1}{c|}{}                         & SAD          & MSE ($10^{-3}$)        & SAD        & MSE ($10^{-3}$)       \\ \shline
Baseline                             & 43.7         & 4.5              & 39.8       & 11.2            \\
Baseline + Deep Supervision                                    & 37.8         & 3.7              & 36.3       & 9.5            \\
Baseline + Fusion Conv                                    & 38.1         & 3.2              & 36.9       & 8.8            \\
PRN w/o CutMask                             & 33.9         & 2.9              & 32.8       & 7.5            \\
PRN                          & 32.3            & 2.5                 & 32.1          & 7.0               \\ \shline
\end{tabular}
\caption{Ablation studies on Composition-1k dataset. Baselines: a ResNet34-UNet with ASPP; Deep supervision: adding side outputs and deep supervisions; Fusion Conv: use convolutions to combine different outputs.}
\label{tab:DIM_Ablation}
\vspace{-1em}
\end{table}

\paragraph{Testing on Composition-1k.} 
The test set consists of 50 unique objects which are composited with 20 background images chosen from Pascal VOC~\cite{everingham2010pascal}, thus providing 1000 test samples in total. 
We note that since these synthetic datasets use PASCAL VOC images as background which may contain other salient objects, saliency/segmentation models may not be applicable to obtain a reasonable coarse mask. To best fairly compare MG Matting with other trimap-based methods, we test our model under two settings: 1) TrimapFG: We adopt the confident foreground regions in a trimap as a coarse guidance mask for our network; 2) Trimap: We normalize trimap to $[0,1]$ with the unknown pixels being $0.5$ and use this soft mask as guidance. We follow the the evaluation setting in Composition-1k which only computes the evaluation on the unknown region.

We summarize the alpha results and foreground color results in Table~\ref{tab:DIM} and Table~\ref{tab:DIM_Color} respectively. We note that although our model is not trained with trimap, it still shows great robustness and transferability on these unseen types of guidance. Our model surpasses previous state-of-the-art models by a large margin. It also performs consistently considering the gap between trimap and trimapFG. We also note that our foreground color prediction not only reduces the errors significantly, but also produces much smoother results (see Fig.~\ref{fig:color_compare}), which is desired in complex real-world scenarios where alpha matte can be noisy.

\paragraph{Testing on Distinction-646.} Distinction-646~\cite{qiao2020attention} is a recent synthetic matting benchmark dataset, which improves the diversity of Composition-1k. It contains 1000 test samples obtained in a similar manner as Composition-1k. However, this dataset is released without official trimaps or other types of guidance, making it difficult to compare with previously reported results.
Therefore, we use this benchmark mainly as a testbed to show how our method can refine a matte produced by another method. 

We test a few state-of-the-art trimap-based baselines trained on Composition-1k. We firstly generate trimaps from ground-truth alpha mattes by thresholding and unknown region is dilated by kernel size 20. Then, we use these trimap-based methods to generate the matting results. Finally, we use these predicted alpha mattes as the guidance to our MG Matting method, and produce refined mattes.

As shown in Table~\ref{tab:Distinction}, using the MG Matting as a refinement method consistently improves the results of other state-of-the-art methods. We also show the results reported by~\cite{qiao2020attention} in Table~\ref{tab:Distinction} for reference.   

\begin{table*}[!t]
 \centering
 \scriptsize
 \setlength{\tabcolsep}{0.0pt}
 \begin{tabular}{ccccccc}
     \includegraphics[trim={0 2.55cm 0 1cm},clip,width=0.142\textwidth]{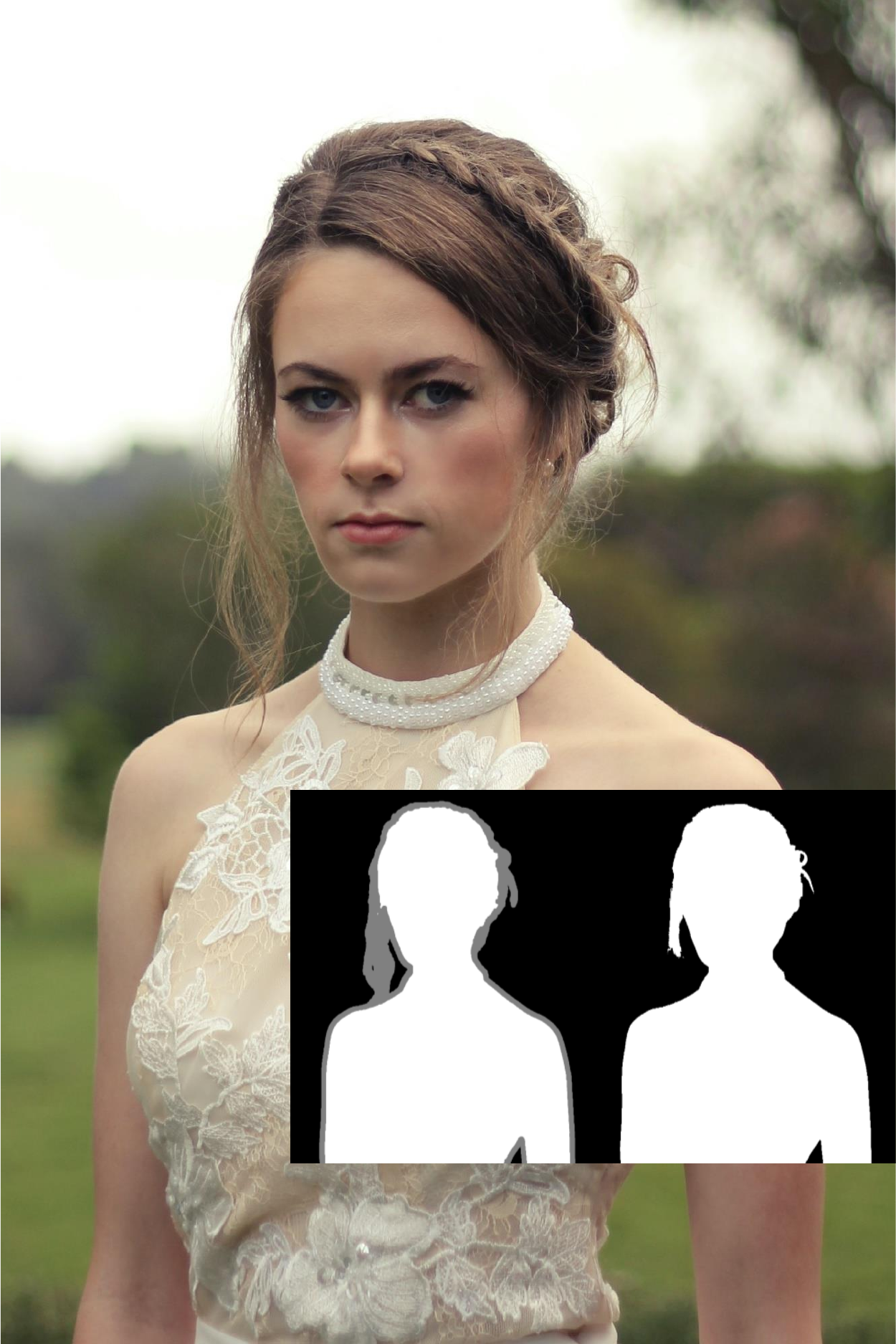} &
     \includegraphics[trim={0 2.55cm 0 1cm},clip,width=0.142\textwidth]{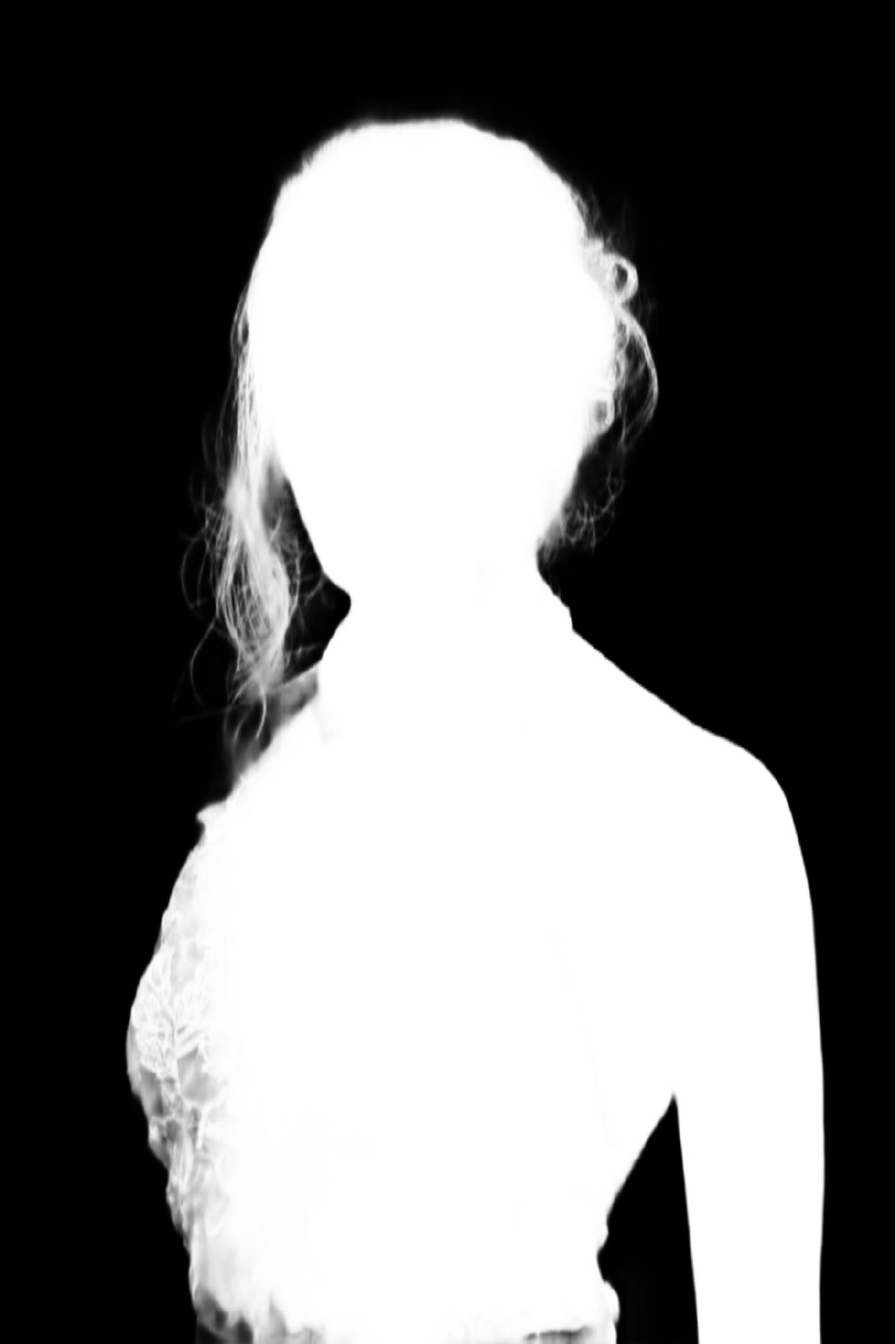} &
     \includegraphics[trim={0 2.55cm 0 1cm},clip,width=0.142\textwidth]{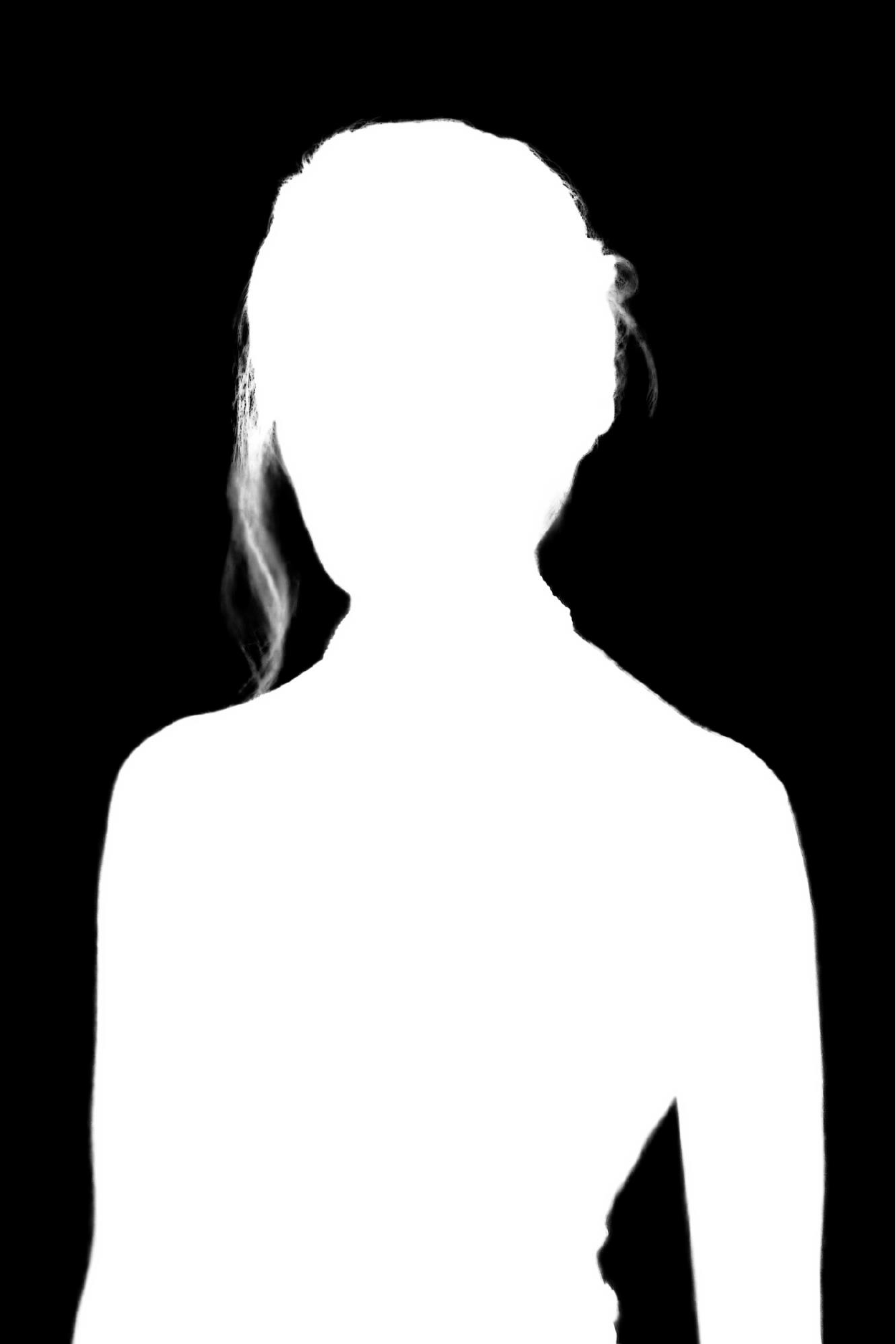} &
     \includegraphics[trim={0 2.55cm 0 1cm},clip,width=0.142\textwidth]{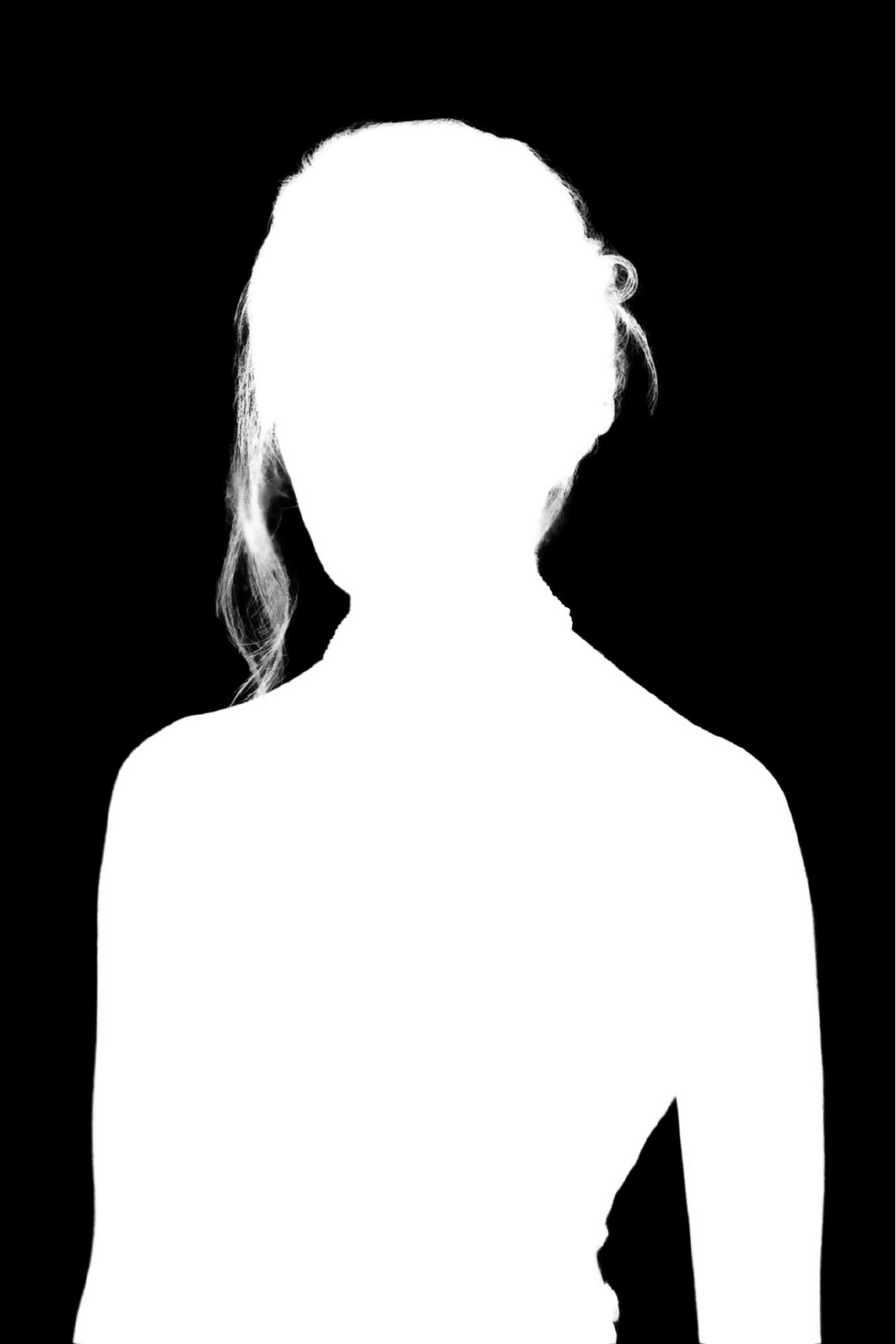} &
     \includegraphics[trim={0 2.55cm 0 1cm},clip,width=0.142\textwidth]{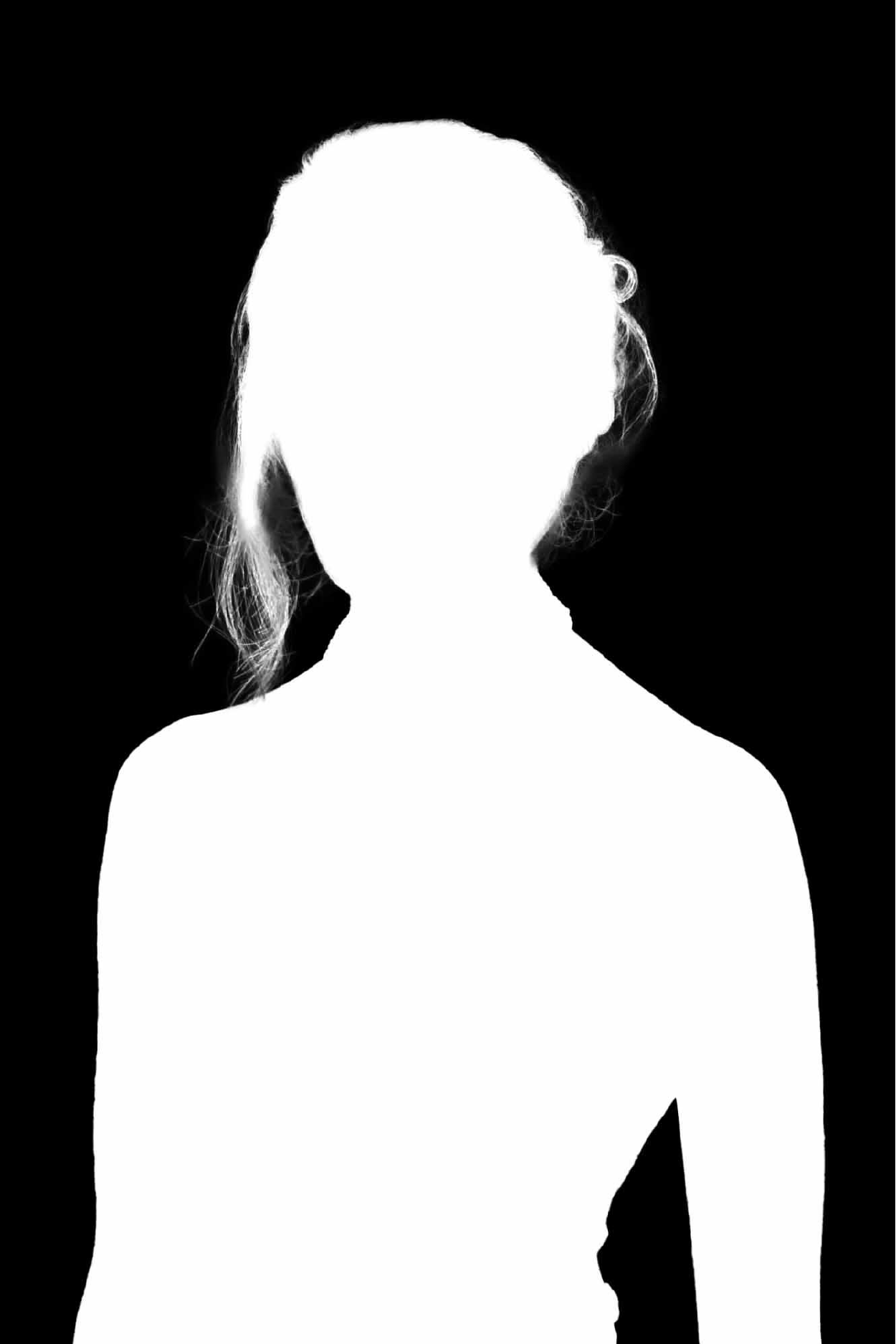} &
     \includegraphics[trim={0 2.55cm 0 1cm},clip,width=0.142\textwidth]{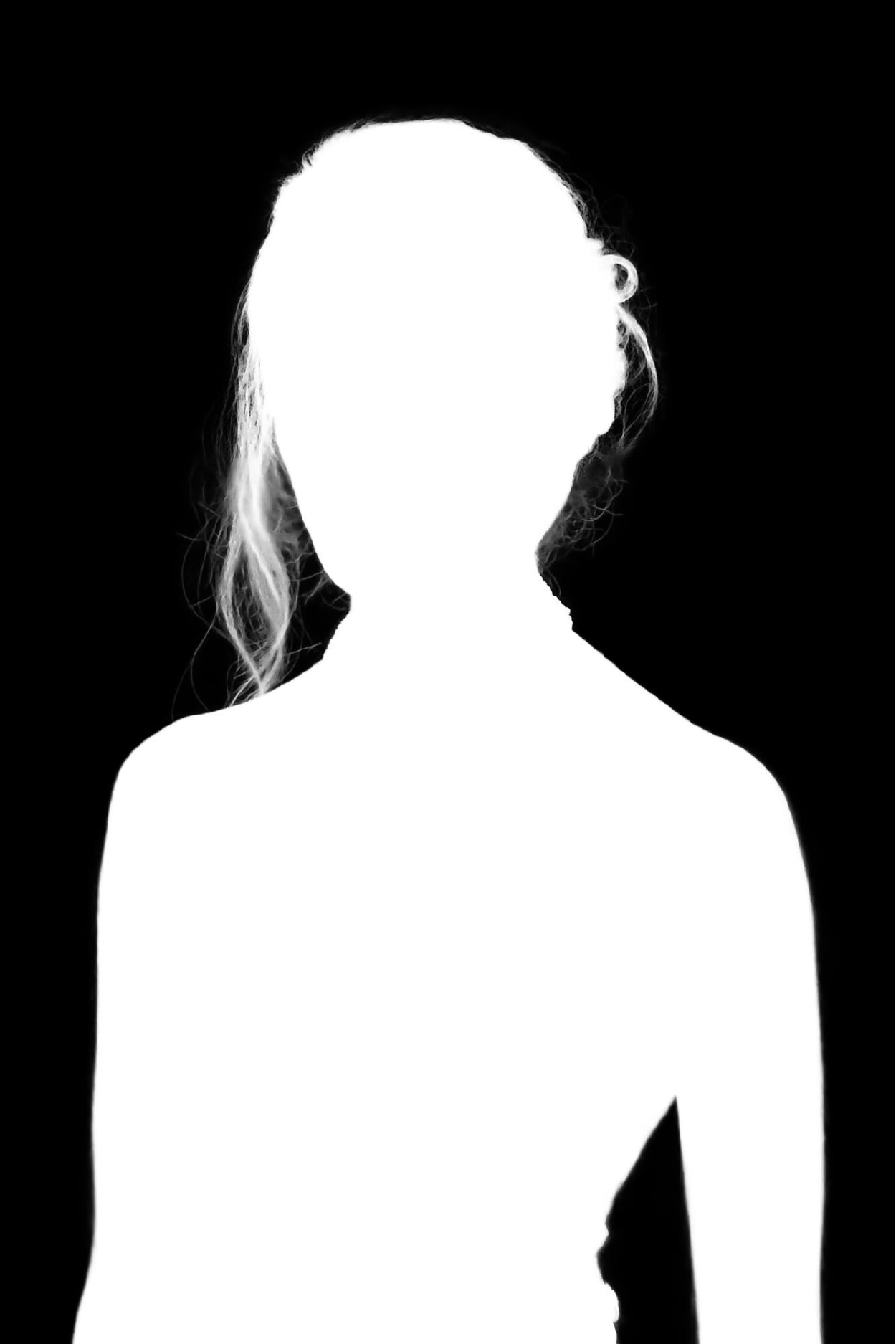} &
     \includegraphics[trim={0 2.55cm 0 1cm},clip,width=0.142\textwidth]{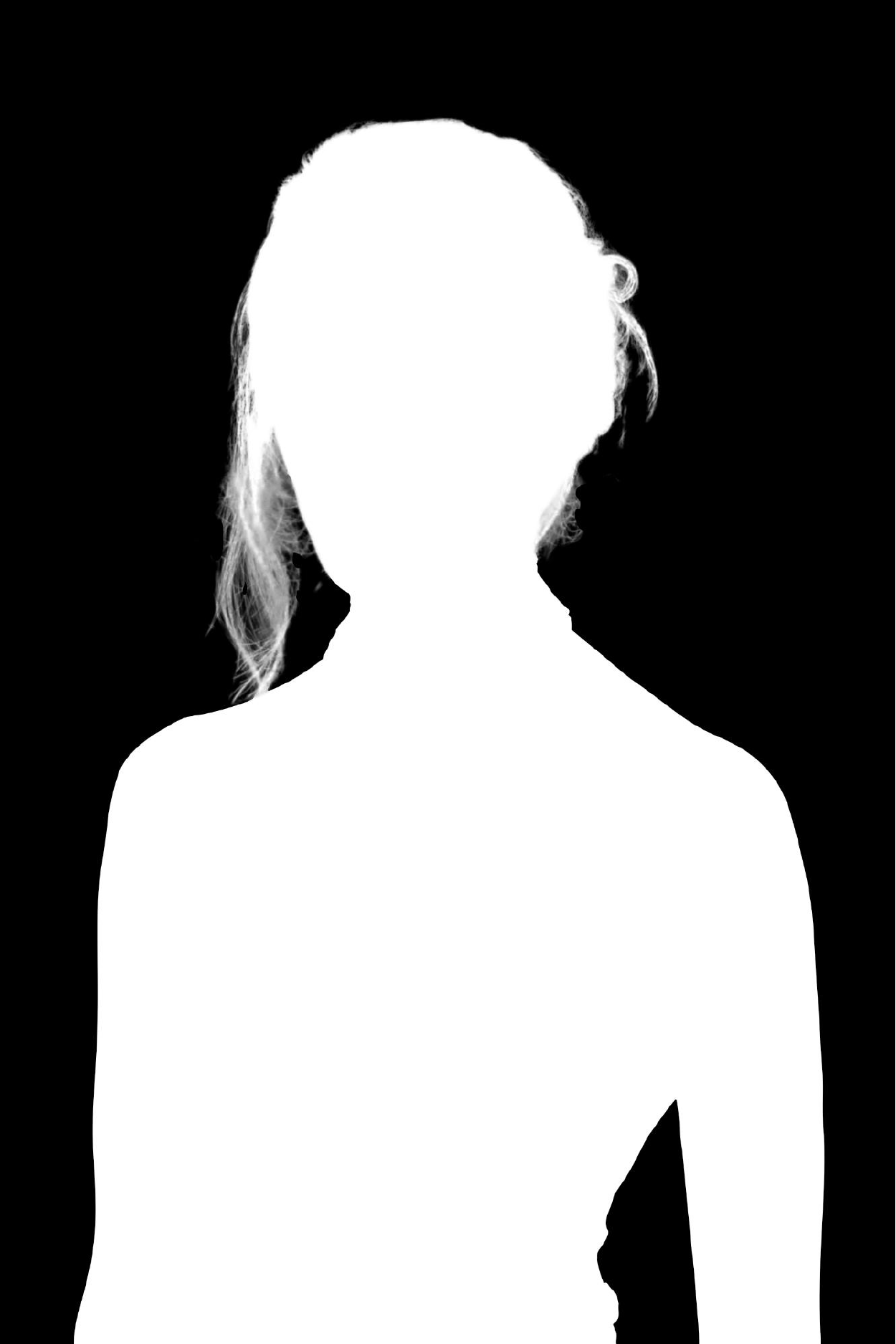}
     \\
     Image w/ guidance &
     LFM~\cite{zhang2019late} &
     GCA~\cite{li2020natural} &
     CA~\cite{hou2019context} &
     PhotoShop &
     MG (Ours) &
     GT
     \\
     \includegraphics[width=0.142\textwidth]{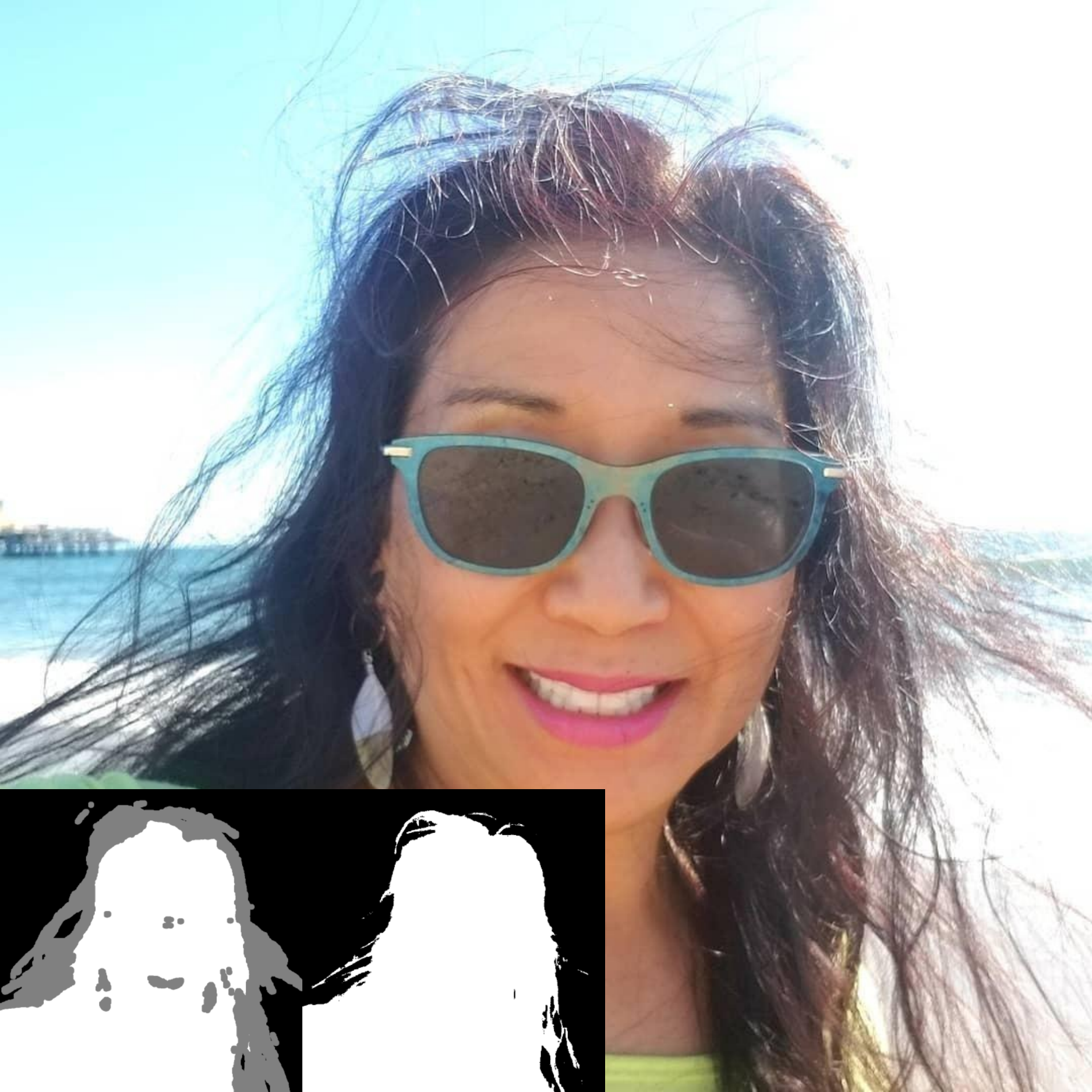} &
     \includegraphics[width=0.142\textwidth]{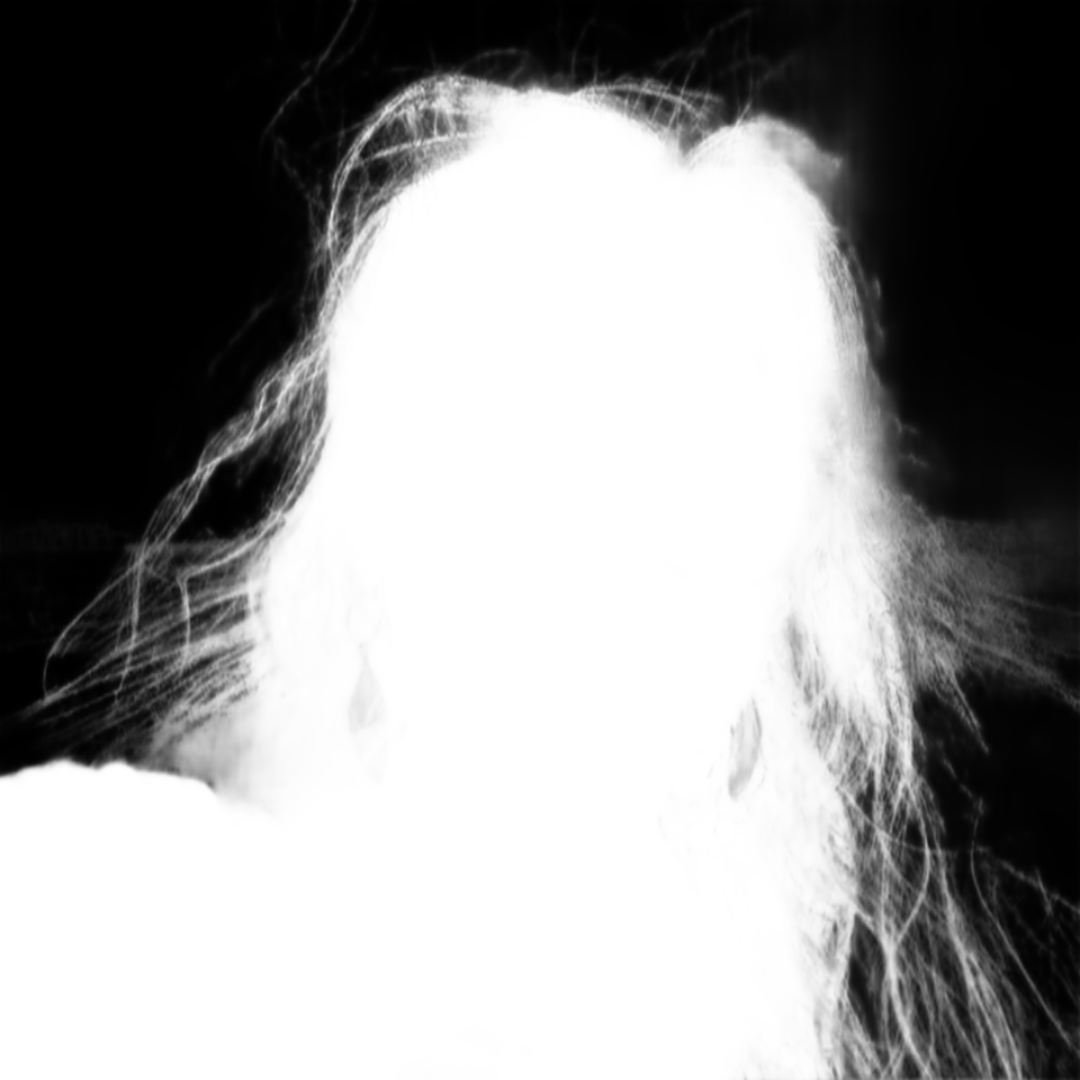} &
     \includegraphics[width=0.142\textwidth]{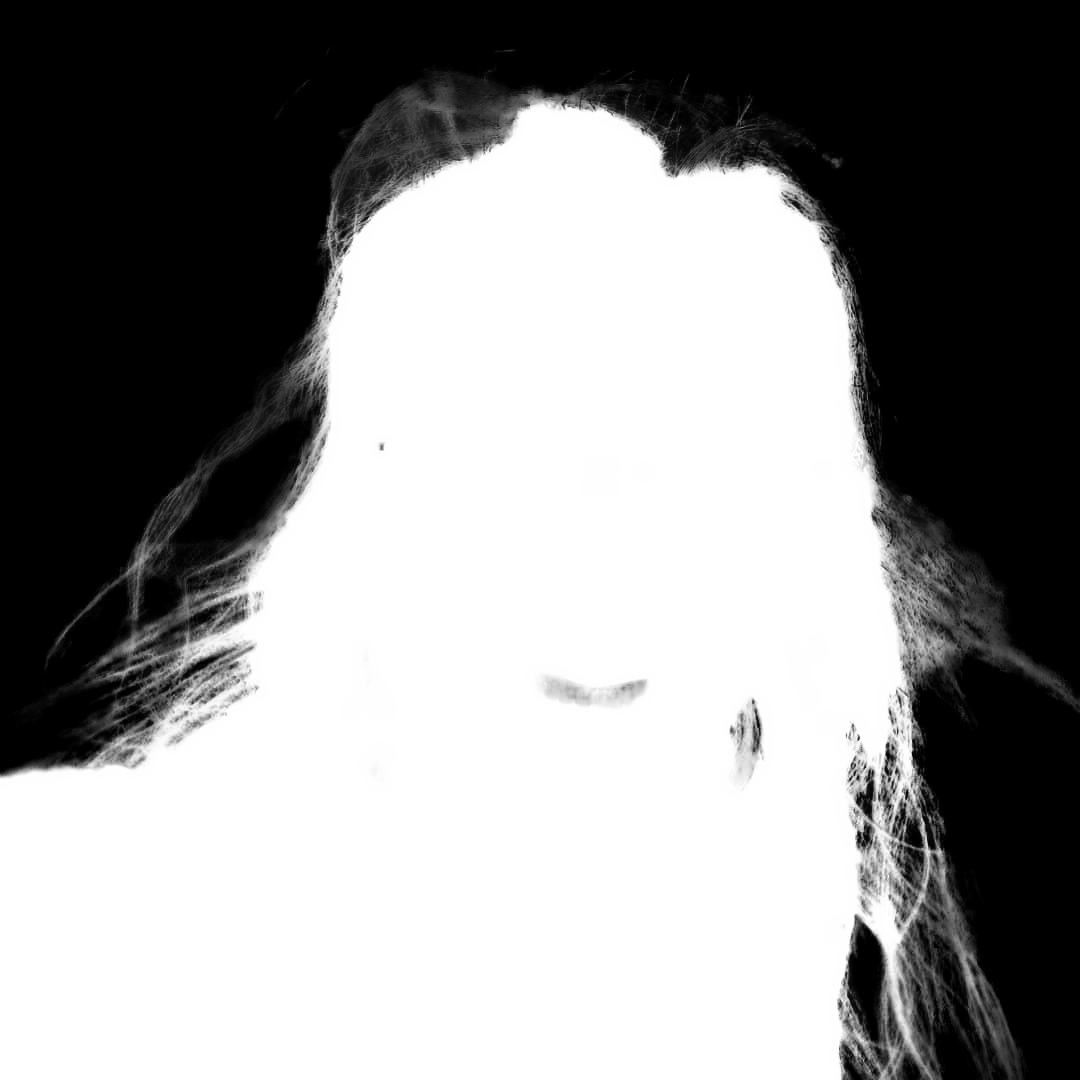} &
     \includegraphics[width=0.142\textwidth]{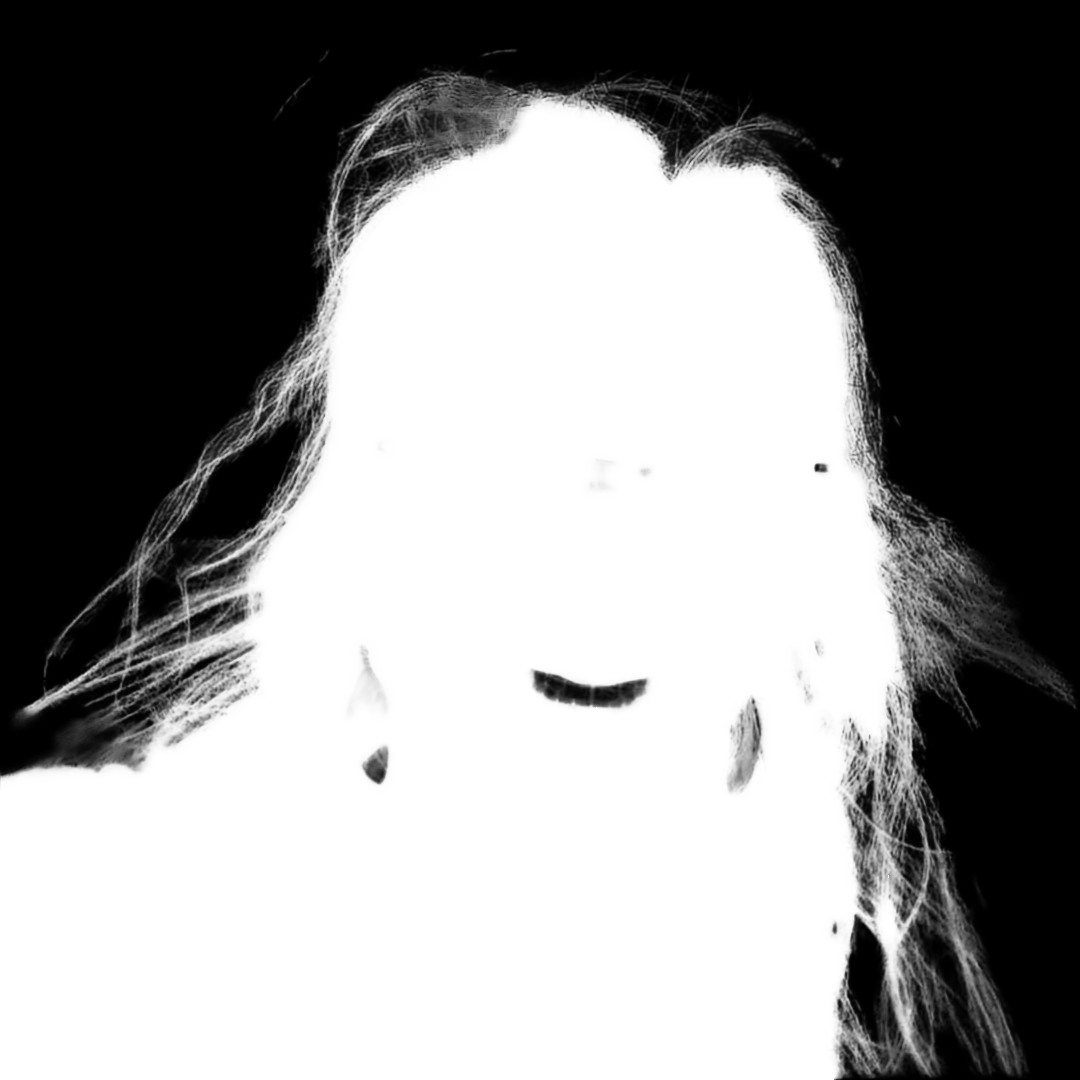} &
     \includegraphics[width=0.142\textwidth]{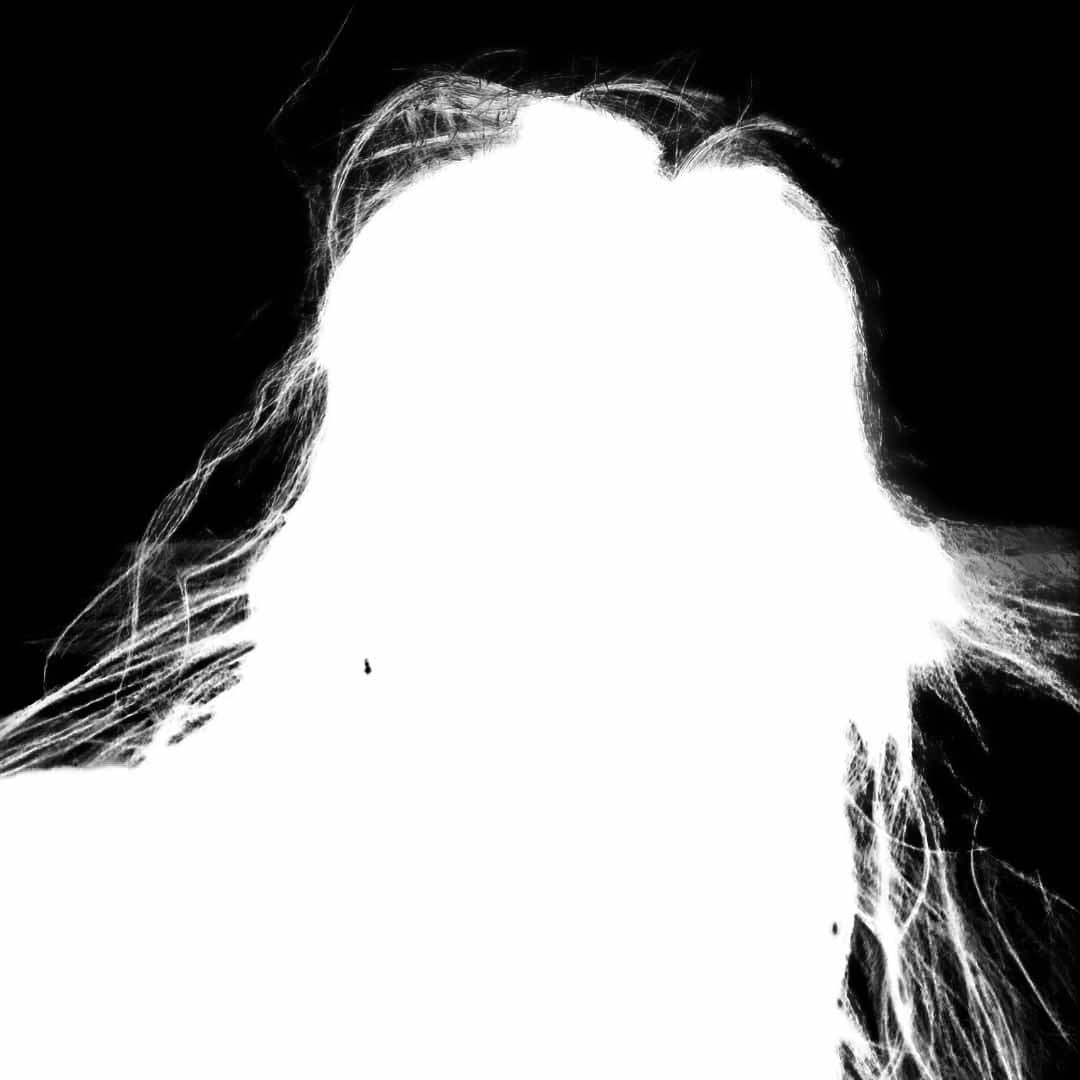} &
     \includegraphics[width=0.142\textwidth]{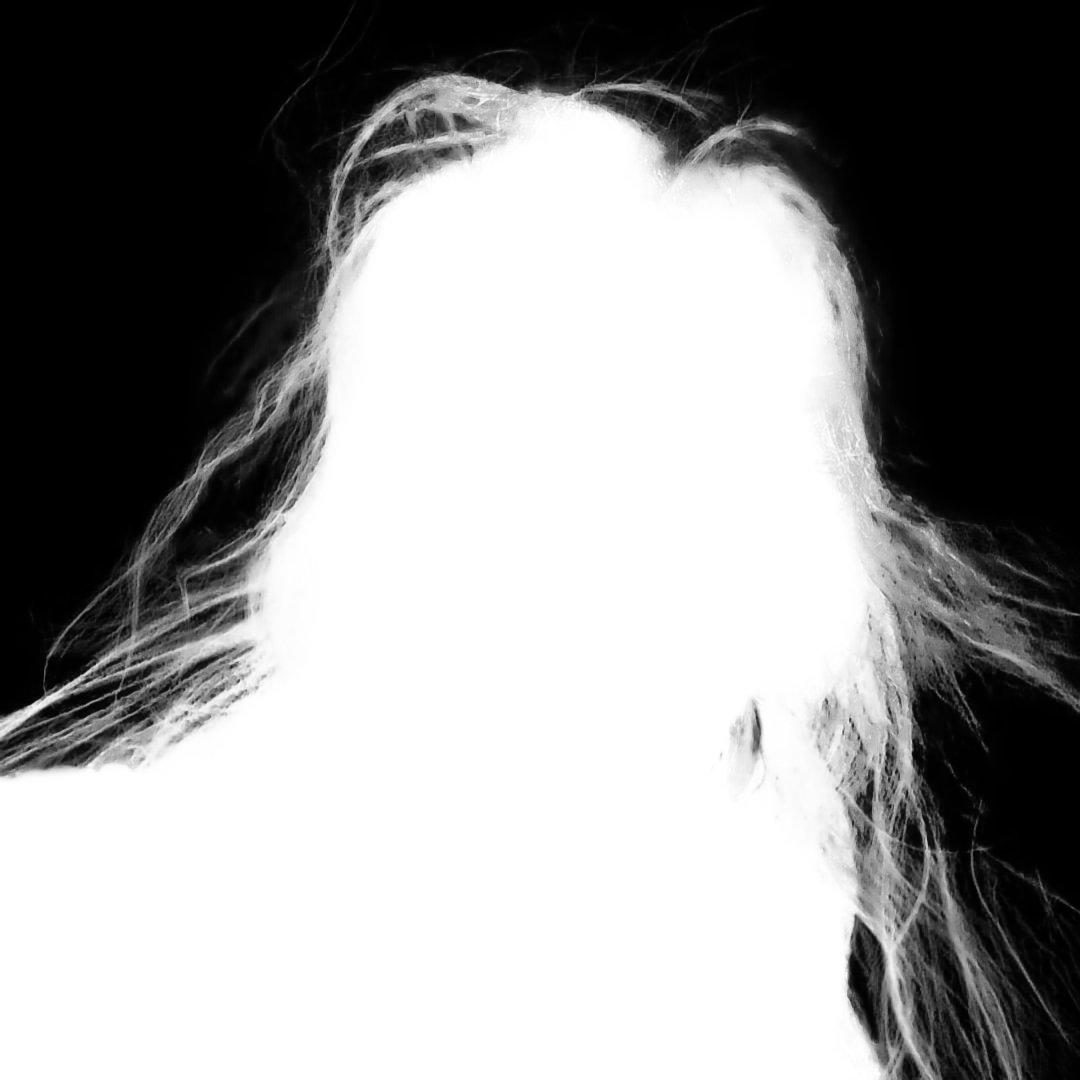} &
     \includegraphics[width=0.142\textwidth]{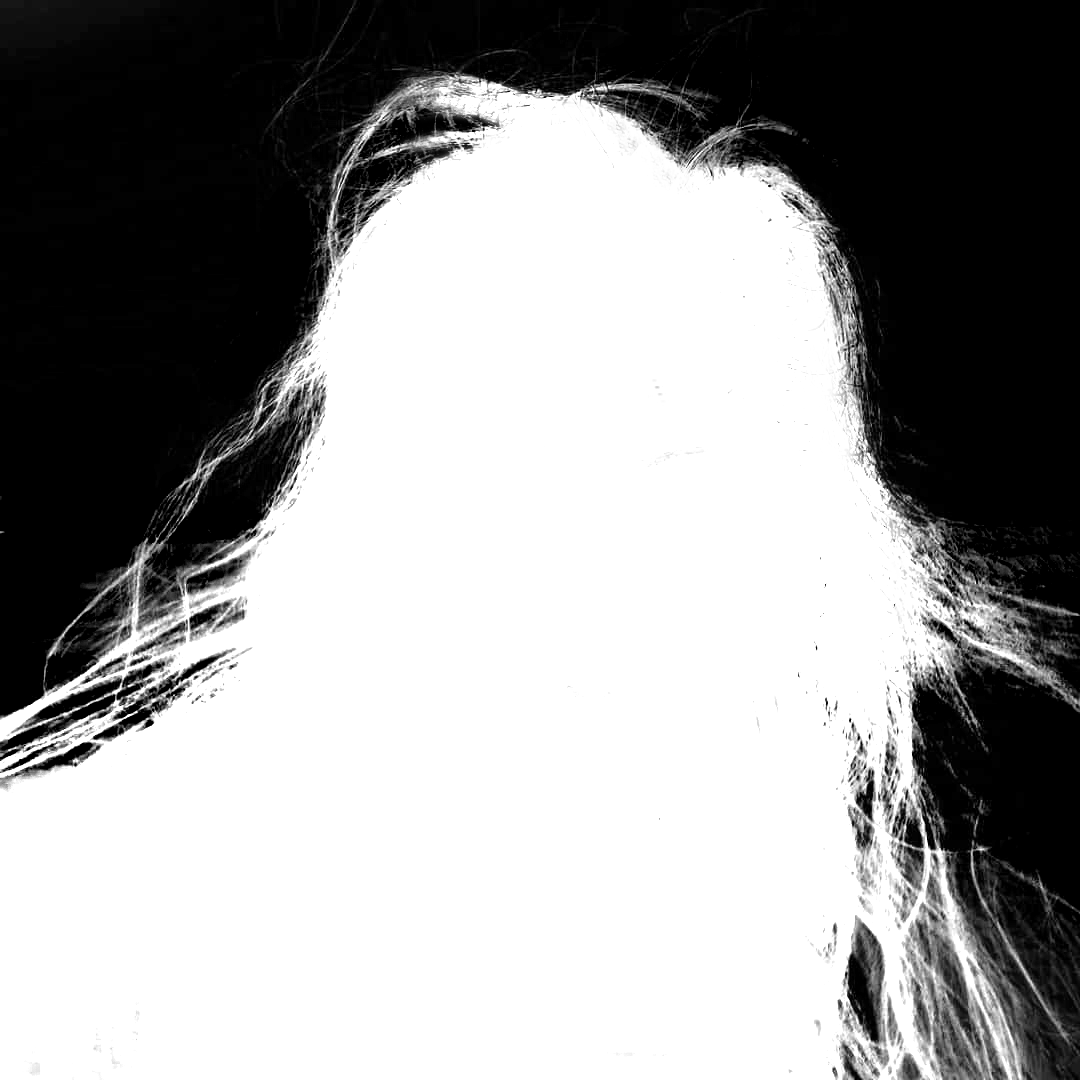}
     \\
     Image w/ guidance &
     LFM~\cite{zhang2019late} &
     GCA~\cite{li2020natural} &
     CA~\cite{hou2019context} &
     PhotoShop &
     MG (Ours) &
     GT
     \\
     \includegraphics[trim={0 5.2cm 0 0},clip,width=0.142\textwidth]{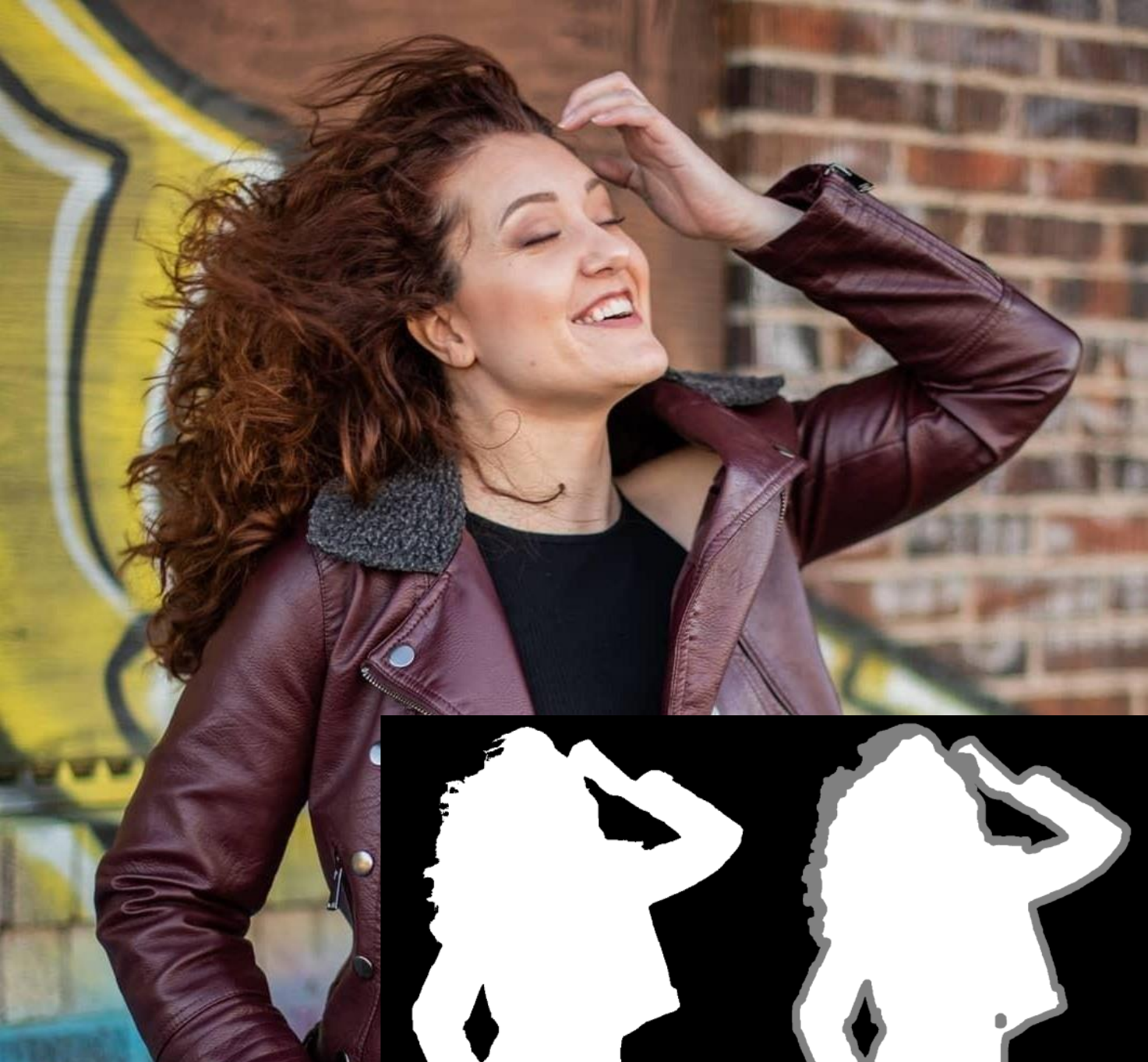} &
     \includegraphics[trim={0 5.2cm 0 0},clip,width=0.142\textwidth]{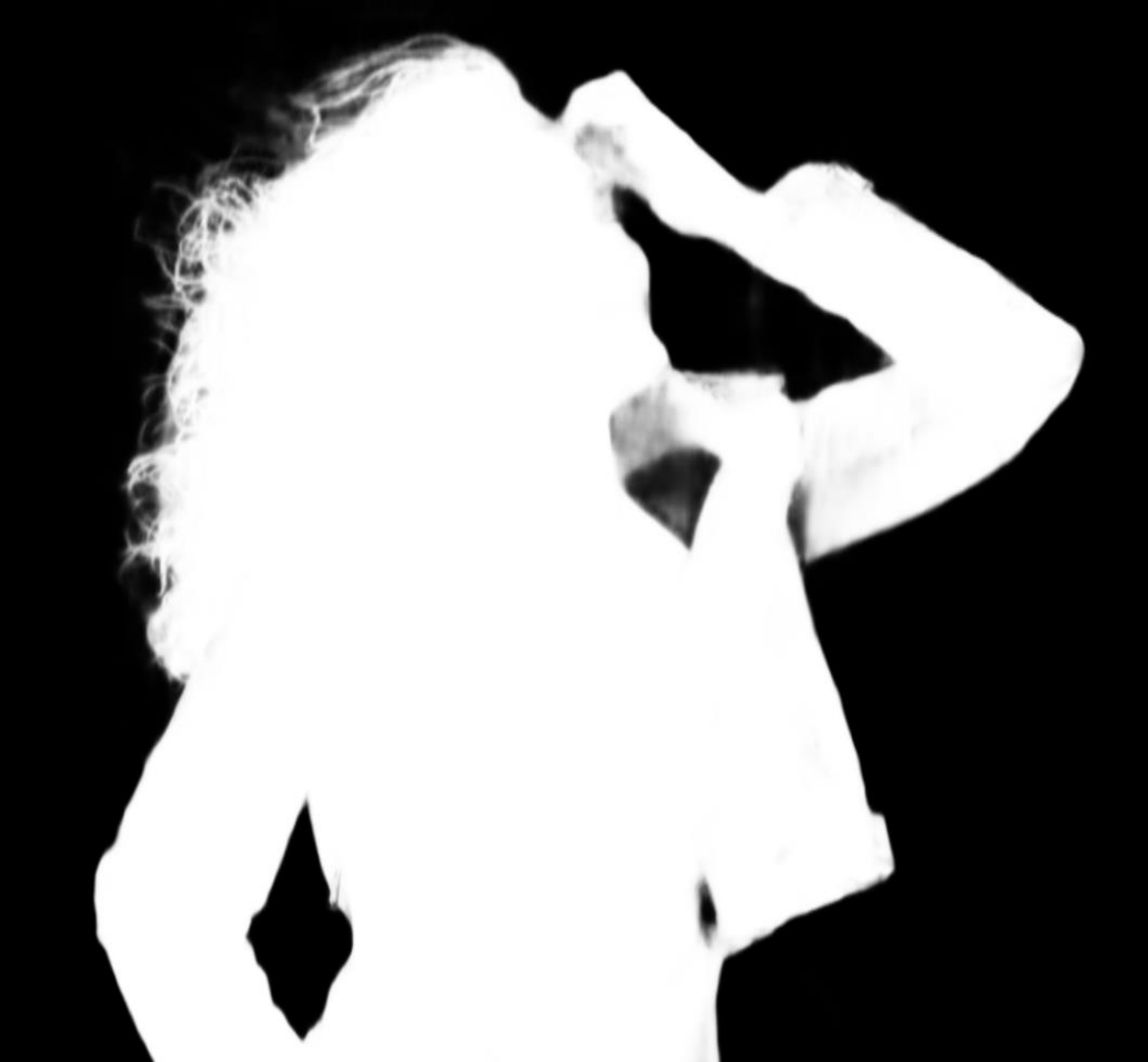} &
     \includegraphics[trim={0 5.2cm 0 0},clip,width=0.142\textwidth]{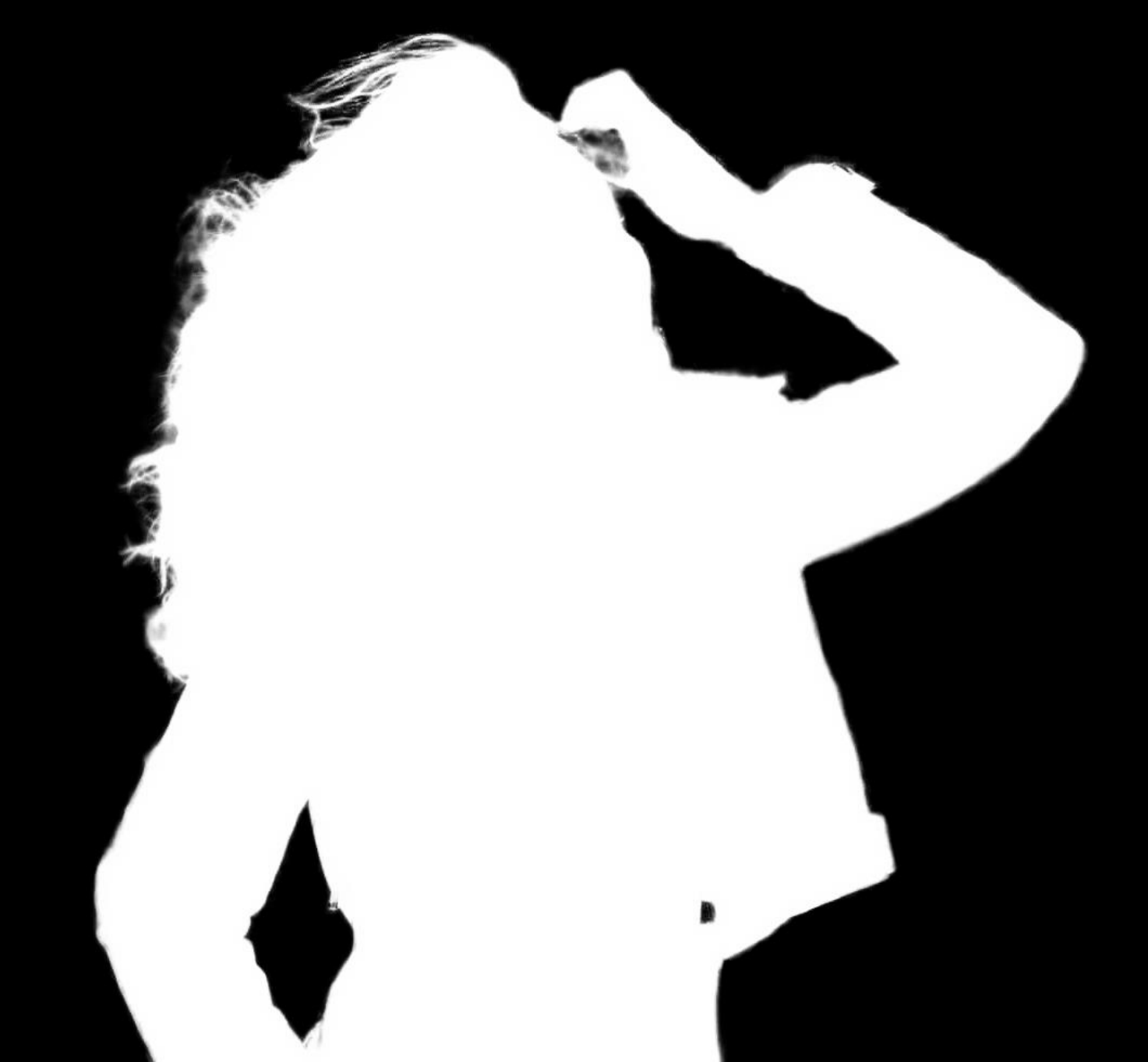} &
     \includegraphics[trim={0 5.2cm 0 0},clip,width=0.142\textwidth]{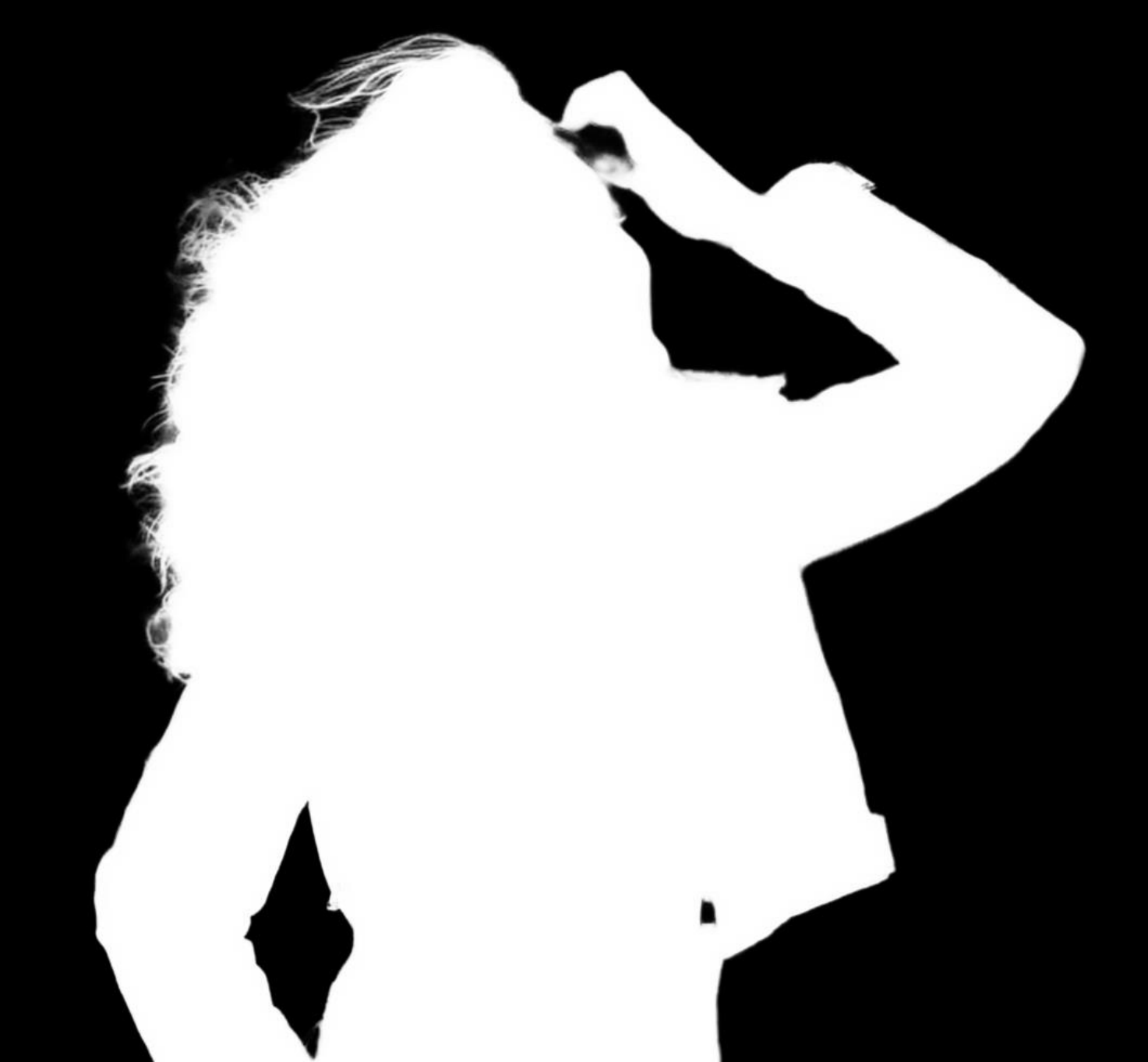} &
     \includegraphics[trim={0 5.2cm 0 0},clip,width=0.142\textwidth]{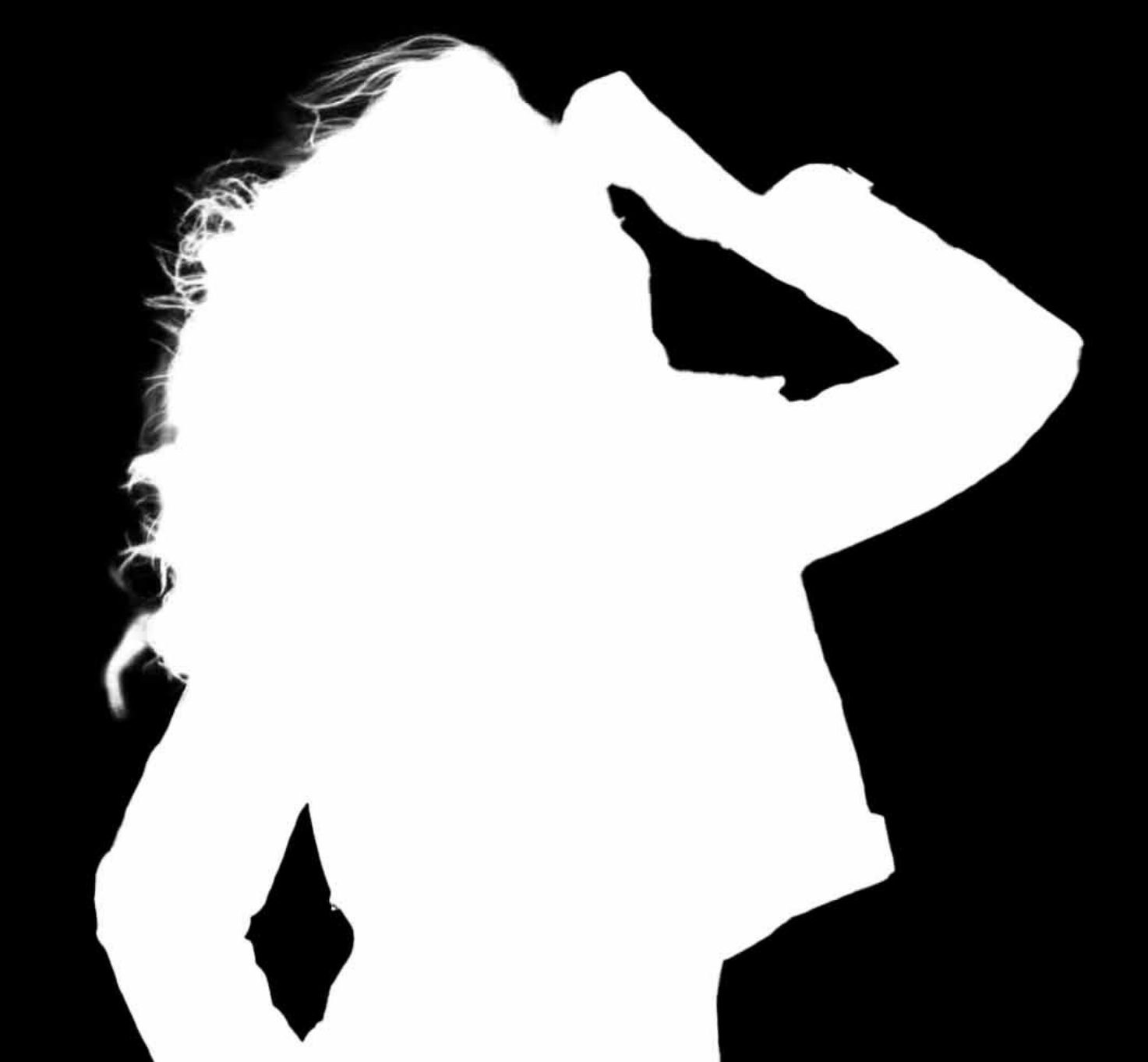} &
     \includegraphics[trim={0 5.2cm 0 0},clip,width=0.142\textwidth]{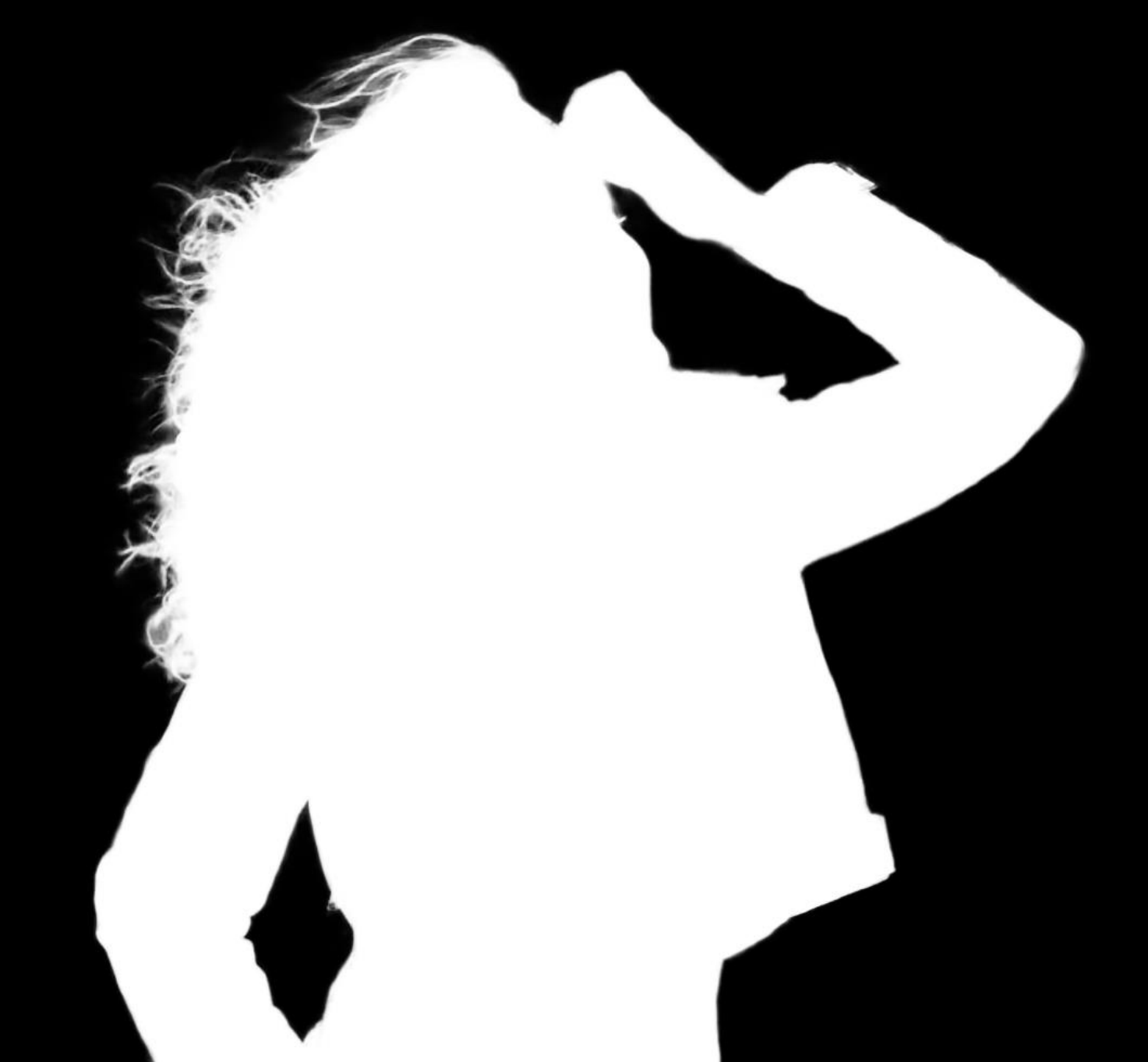} &
     \includegraphics[trim={0 5.2cm 0 0},clip,width=0.142\textwidth]{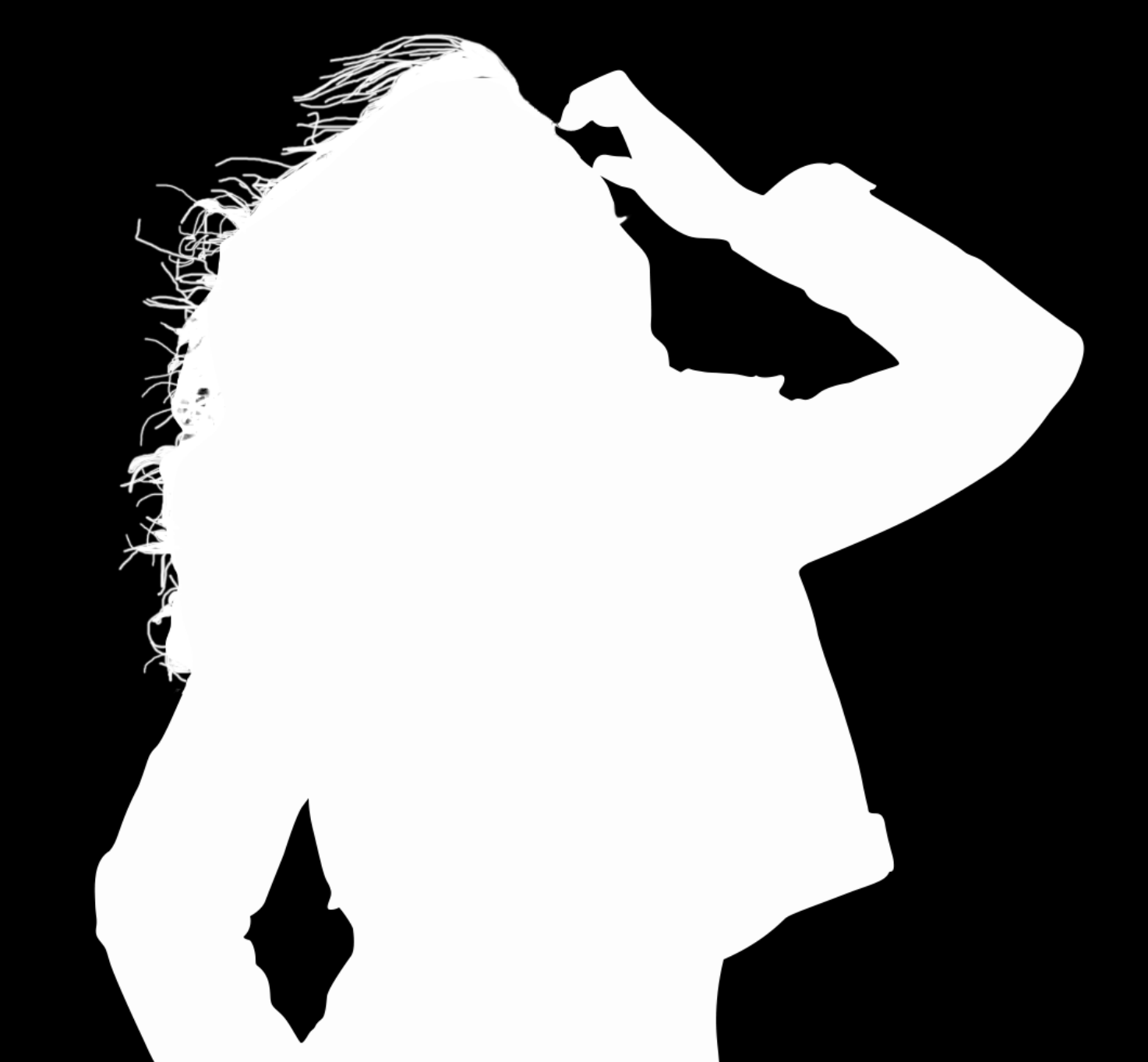}
     \\
     Image w/ guidance &
     LFM~\cite{zhang2019late} &
     GCA~\cite{li2020natural} &
     CA~\cite{hou2019context} &
     PhotoShop &
     MG (Ours) &
     GT
 \end{tabular}
 \captionof{figure}{The visual comparison results among different methods on our portrait test set. We visualize representative examples with both high-quality studio-level portraits and selfies with strong noises. MG Mating performs well on different quality images and can maintain details. We note that our results, though only trained on composition-1k, are not only superior to previous state-of-the-art but also produces comparable or better results than commercial methods in PhotoShop.}
 \label{fig:FinalVis}
 \vspace{-1em}
\end{table*}

\paragraph{Ablation Studies.}
To validate the design of PRN and the introduced guidance perturbation, we conduct ablations studies as summarized in Table~\ref{tab:DIM_Ablation}. Trimap is used as guidance masks in these experiments. However, we do not assume that the guidance type is known, so we purposefully do not use it to post-process the prediction by replacing the known foreground and background region. Instead,  we report the two scores calculated over the whole image and the unknown region respectively for a more comprehensive evaluation of the robustness of our method.

We report ablations of different variants in Table~\ref{tab:DIM_Ablation}. Baseline refers to a pure backbone without any add-ons. Adding side outputs and deep supervision to baseline improves the performance on both whole image or unknown area. We also try to use two convolution layers to fuse different outputs. However, linearly fusing the side outputs may not lead to better results. In contrast, the proposed PRN can better coordinate the semantic refinement and low-level detail refinement at different levels, thus obtaining a consistent improvement. We also show that the CutMask perturbation can further improve both the performance and robustness.

We also validate the effectiveness of RAB. We calculate the MSE and SAD of foreground color ($\mathbf{F}$) over foreground regions (\ie $\alpha>0$). The baseline achieves $\mathrm{MSE}=0.00623$ and $\mathrm{SAD}= 82.30$, while with RAB, the performance is boosted to $\mathrm{MSE}=0.00321$ and $\mathrm{SAD}=62.01$.

\begin{table}[tb]
\setlength{\tabcolsep}{1.0pt}
\centering
\footnotesize
\begin{tabular}{L{0.5\linewidth}|C{0.11\linewidth}C{0.11\linewidth}|C{0.11\linewidth}C{0.11\linewidth}}
\shline
\multicolumn{1}{c|}{\multirow{3}{*}{Methods}} & \multicolumn{2}{c|}{Whole Image} & \multicolumn{2}{c}{Details} \\ \cline{2-5} 
\multicolumn{1}{c|}{}                         & SAD          & MSE ($10^{-3}$)        & SAD        & MSE ($10^{-3}$)       \\ \shline
Deep Image Matting~\cite{xu2017deep}                             & 28.5         & 11.7              & 19.1       & 74.6            \\ 
GCA Matting~\cite{li2020natural}                                    & 29.2         & 12.7              & 19.7       & 82.3            \\ 
IndexNet Matting~\cite{lu2019indices}                               & 28.5         & 11.5              & 18.8       & 72.7            \\ 
Context-Aware Matting~\cite{hou2019context}                          & 27.4            & 10.7                 & 18.2          & 66.2               \\ 
Late Fusion Matting~\cite{zhang2019late}                            & 78.6            & 39.8                 & 24.2          & 88.3               \\ \shline
Ours                               & \textbf{26.8}         & \textbf{9.3}               & \textbf{17.4}       & \textbf{55.1}            \\ \shline
\end{tabular}
\caption{Results on Real-world Portrait test set.}
\label{tab:portrait}
\vspace{-1em}
\end{table}

\section{Experiments on Real-world Portrait Dataset}
\label{Experiments_portrait}
We note that although the synthetic datasets are well-established benchmarks and provide sufficient data to train a good model, it remains an open question whether models trained on them are robust enough and can produce comparable results in real images. For example,~\cite{hou2019context} found that some easy data augmentations such as re-JPEGing and gaussian blur can avoid some shortcomings of the synthetic dataset and significantly improve the model's performance on real-world images, though at a cost of higher errors on the synthetic benchmark. This begs the question: \emph{can the results on synthetic matting dataset reflect the performance on real images?}

Evaluation on real-world images is thus very crucial. However, due to the lack of high-quality matting benchmark datasets of real images, most previous models mainly compare their matting results visually or through user study.
To better evaluate the matting methods in a real-world scenario, we collect a real-world image matting dataset consisting of $637$ diverse and high-resolution images with matting annotation made by experts. The images in our dataset have various image quality and subjects of diverse poses. Moreover, since the dataset mainly contains solid objects where the main body can be easy to predicted, we also labeled detail masks covering the hair region and other soft tissues, which tells where the most important details of the image are located. By calculating errors in these regions, we can further compare the ability to capture object details for different models.
We will release this dataset for better benchmarking matting methods on real images.

\begin{center}
 \centering
 \footnotesize
 \setlength{\tabcolsep}{0.0pt}
 \begin{table}[t]
 \begin{tabular}{ccc}
    \includegraphics[trim={0 5.5cm 0 3cm},clip,width=0.155\textwidth]{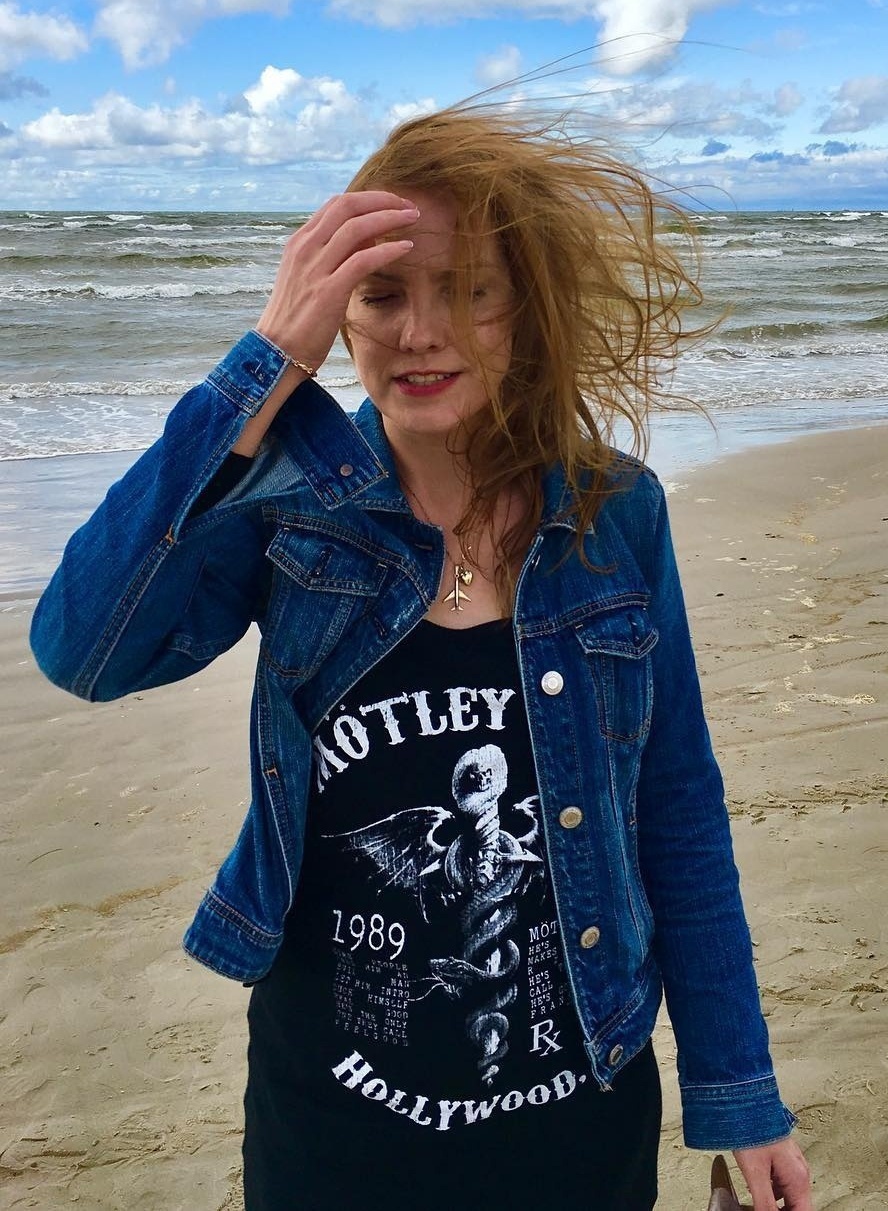} &
     \includegraphics[trim={0 5.5cm 0 3cm},clip,width=0.155\textwidth]{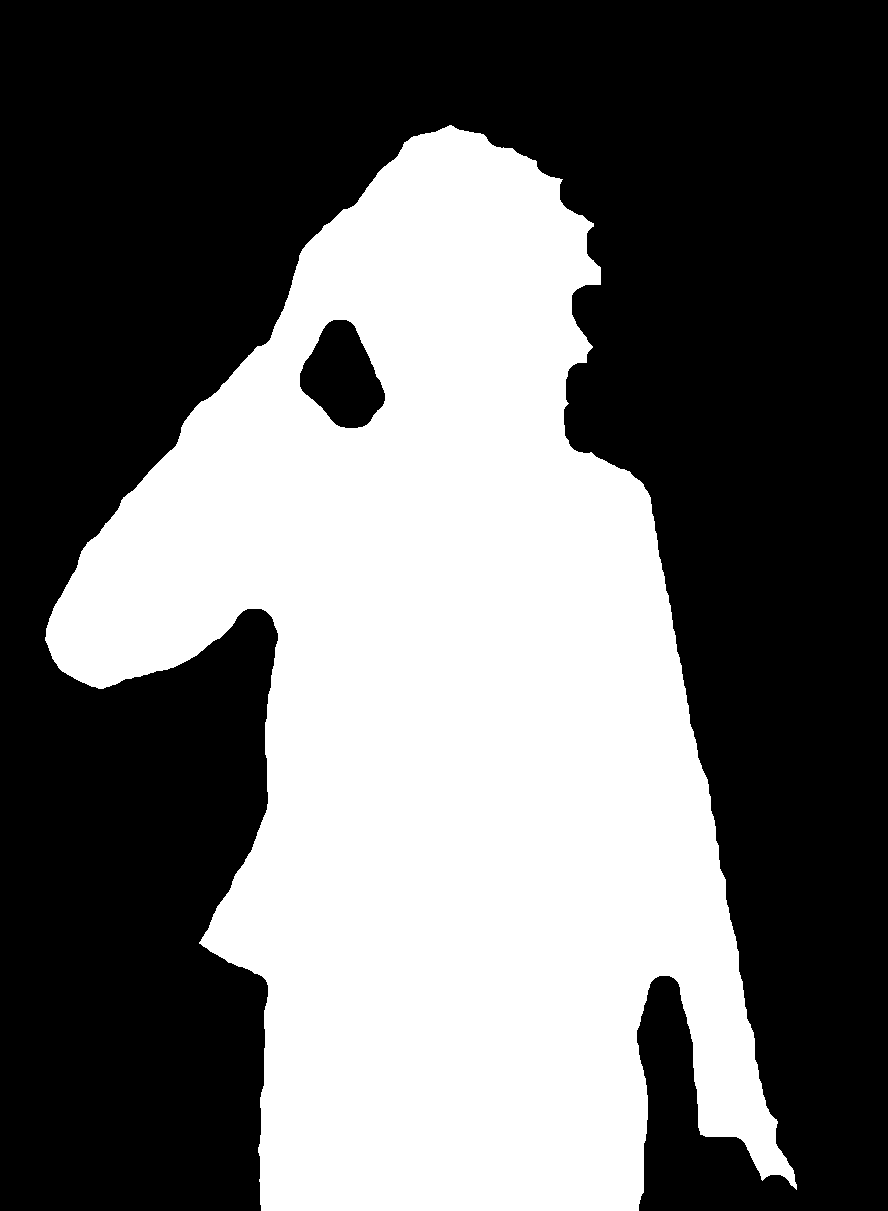} &
     \includegraphics[trim={0 5.5cm 0 3cm},clip,width=0.155\textwidth]{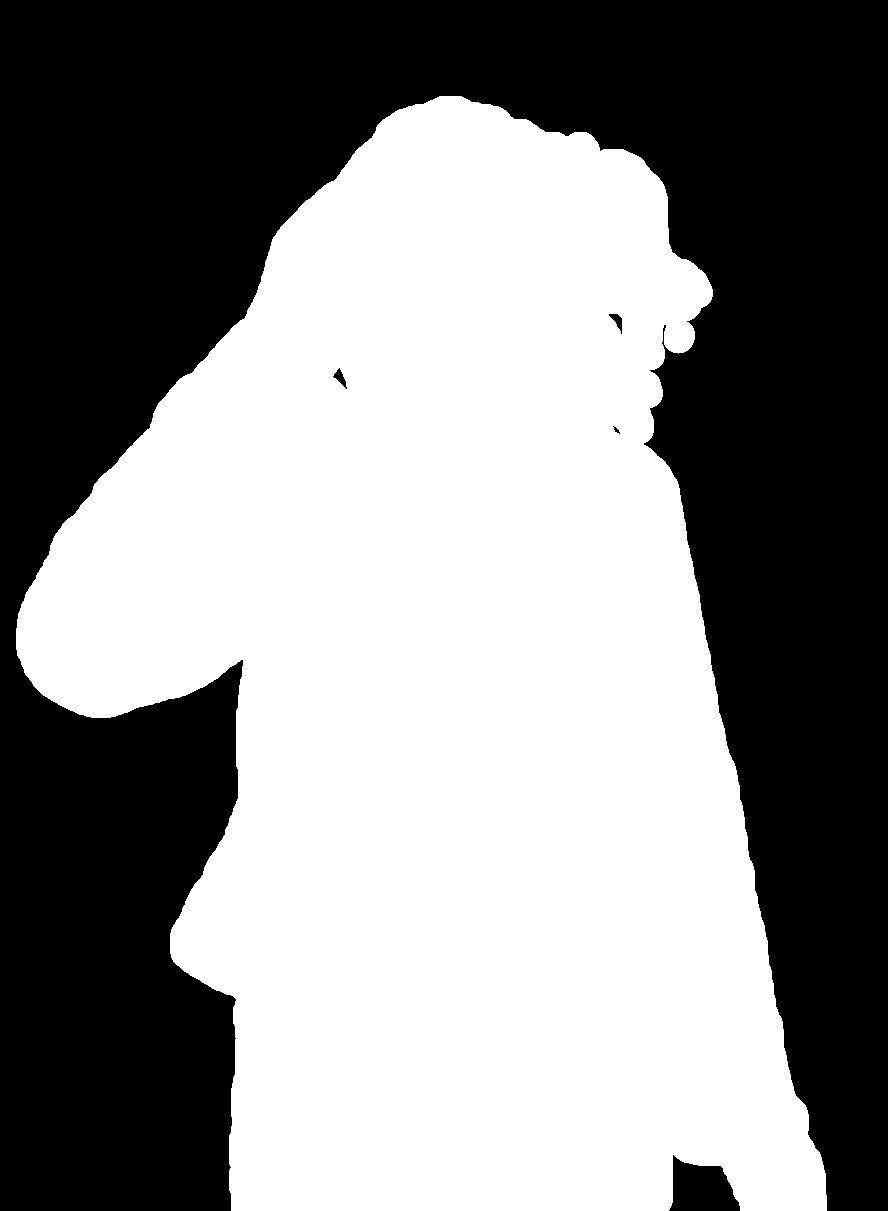}
     \\
     Image &
     Mask-Erode30 &
     Mask-Dilate30
     \\
    \includegraphics[trim={0 5.5cm 0 3cm},clip,width=0.155\textwidth]{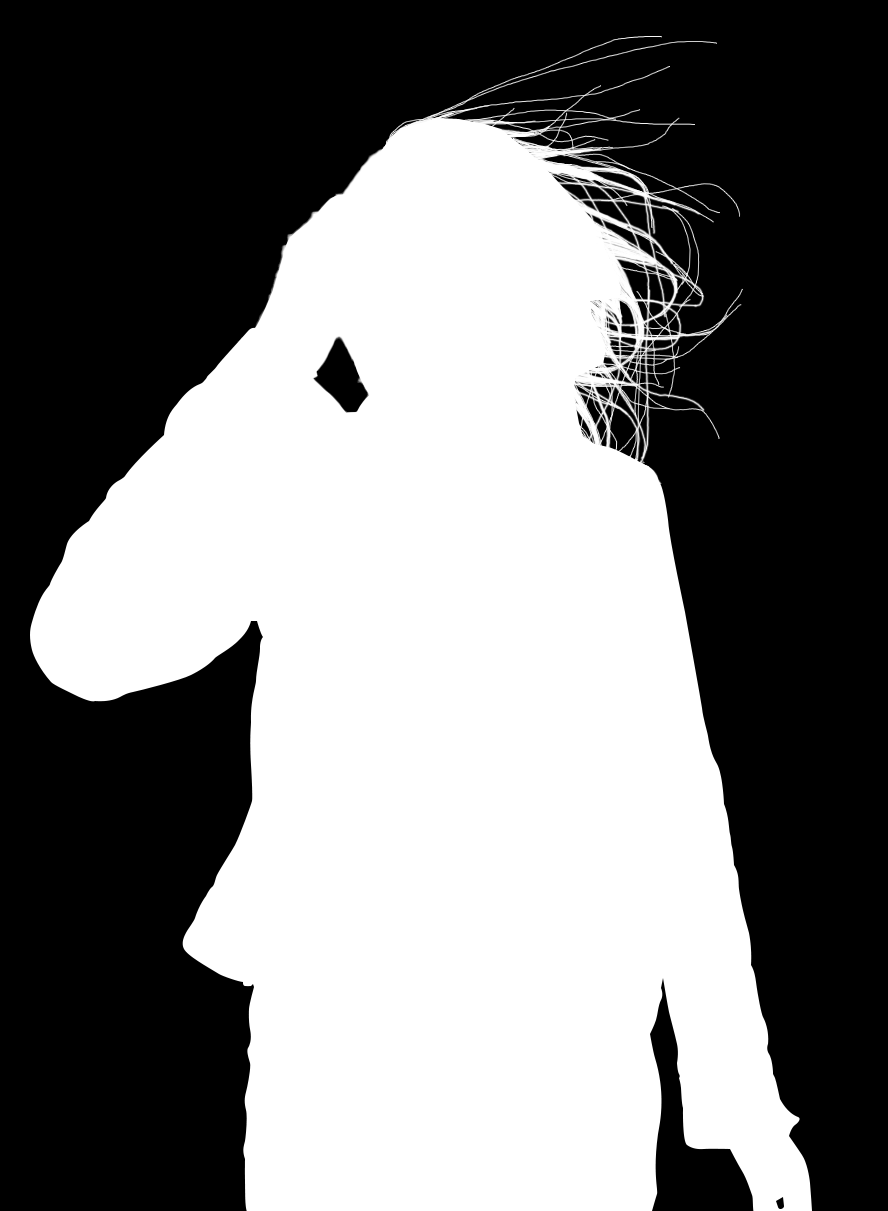} &
     \includegraphics[trim={0 5.5cm 0 3cm},clip,width=0.155\textwidth]{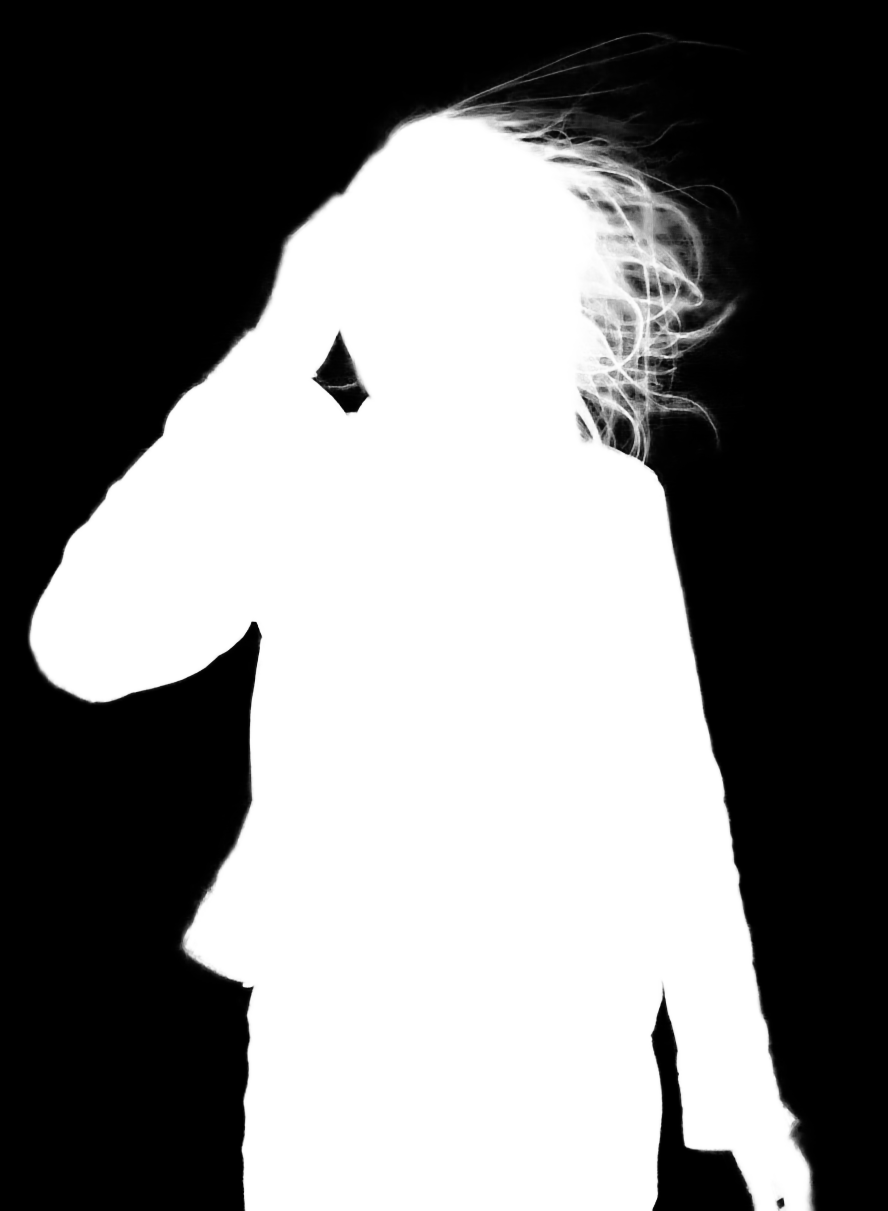} &
     \includegraphics[trim={0 5.5cm 0 3cm},clip,width=0.155\textwidth]{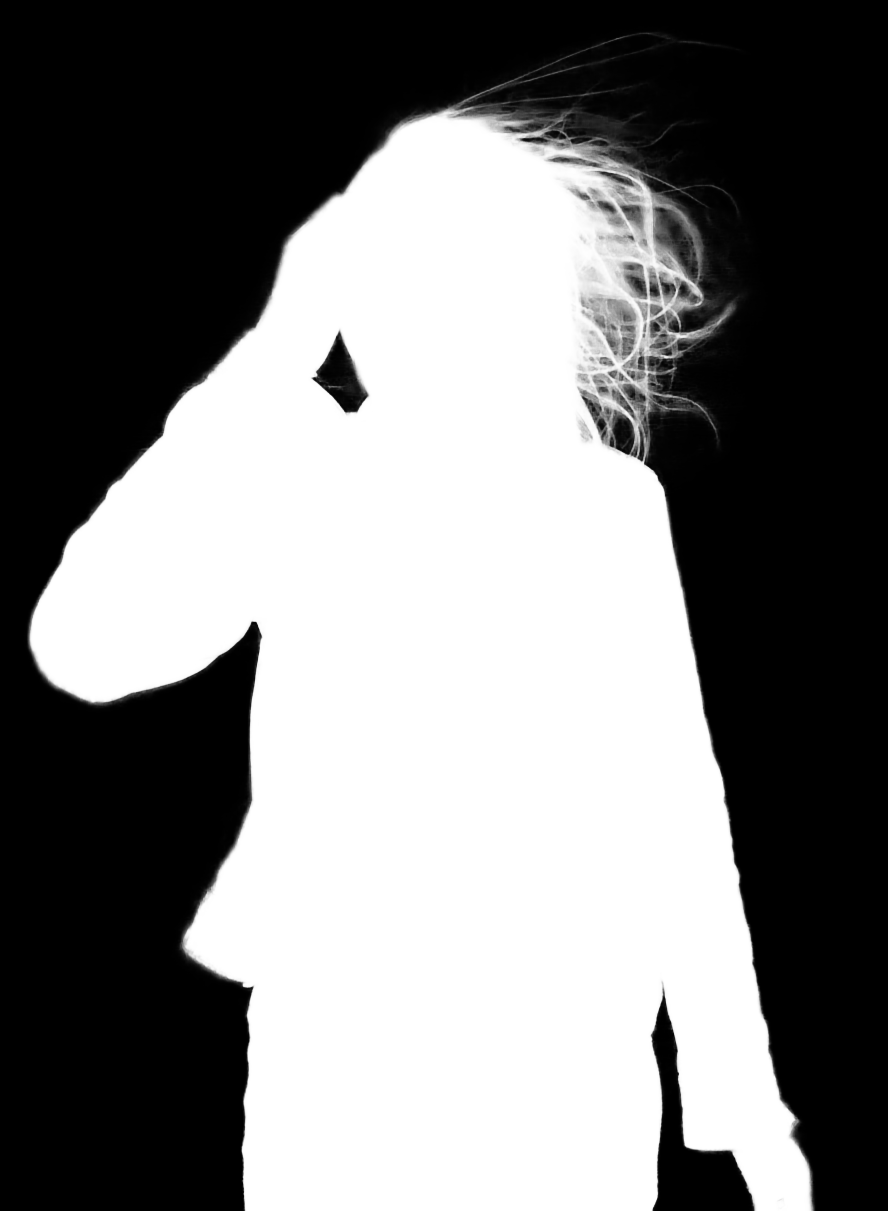} 
     \\
     Ground-Truth &
     Pred-Erode30 &
     Pred-Dilate30
 \end{tabular}
 \captionof{figure}{Our model is robust given different quality guidance masks and produces consistent alpha estimation.}
 \label{fig:robust_vis}
 \end{table}
 \vspace{-1em}
\end{center}

\paragraph{Implementation Details.}
We use the Composition-1k training set to train the model. Considering the semantic gap between the two datasets, we remove the transparent objects from the training data using the data list of~\cite{sengupta2020background}.
Following~\cite{hou2019context}, we also apply re-JEPGing, gaussian blur, and gaussian noises to the input image to make the model better adapt to real-world noises which are rarely seen in the synthetic dataset. Since these augmentations can change the color of the composited training image, thus the original color label may not be applicable. Therefore, we remove the composition loss from the supervision. Other training settings remain the same as in Sec.~\ref{Experiments_DIM}.

For trimap-based baselines, we follow~\cite{sengupta2020background} to generate trimaps from segmentation~\cite{zhang2020deep} automatically by labeling each pixel with foreground class probability $>0.95$ as foreground, $<0.05$ as background, and the rest as unknown, the unknown region is further dilated by $k=20$ to ensure it will not miss the long hairs. For our model, we threshold the segmentation at prob $=0.5$ to a binary mask.

\paragraph{Results.}
We compare the results with state-of-the-art trimap-based methods DIM~\cite{xu2017deep}, GCA~\cite{li2020natural}, IndexNet~\cite{lu2019indices}, Context-Aware Matting~\cite{hou2019context}, and trimap-free method Late Fusion Matting~\cite{zhang2019late} which is trained on Composition-1k training set and an additional portrait dataset. The results of baselines are obtained through either the open-source inference demos or the provided pre-trained weights.

We summarize the results in Table~\ref{tab:portrait} under two settings: Whole Image, where the errors are calculated across the whole image, which can measure the overall quality; Details, where the errors are calculated only in manual-labeled regions containing hair details or other soft areas.

Compared to other methods, our model achieves a superior performance, especially regarding to the detail part, which illustrates its ability to capture the boundary details. We also note that the trimap-free method LFM performs badly, which could be caused by the fact that their portrait training data is not diverse enough and thus limits the the generalizability of their model (see Fig.~\ref{fig:FinalVis} for examples). 

We compare our results with another trimap-free method BSHM~\cite{liu2020boosting}. We contacted the authors and obtained the test results on a 100 images subset of our portrait dataset. Since~\cite{liu2020boosting} can only deal with low-resolution images, we downsample images to longer-side $720$, and the metrics are also computed on this scale.~\cite{liu2020boosting} achieves MSE $0.0155$ and SAD $10.66$ for whole image and MSE $0.0910$ and SAD $7.60$ for detail regions, while our MG Matting obtains a superior performance with MSE $0.0095$ and SAD $8.01$ for whole image and MSE $0.0637$ and SAD $5.94$ for details.

\vspace{0.5ex}
\noindent \textbf{Robustness to Guidance.} To verify how robust our model is to the external guidance mask, we conduct an experiments to feed the network with perturbed external guidance mask. Particularly, we erode/dilate the mask with kernel size $10$, $20$, $30$ respectively. We note that the model predict consistently given differently perturbed external guidance. The SAD error increases from $26.8$ to $27.1$, $27.2$, $27.4$ with mask eroded by $10$, $20$, and $30$ respectively. For dilation, the SAD error goes to $27.0$, $27.4$, $28.1$ with kernel $10$, $20$, $30$ respectively. A visual example is provided in Fig.~\ref{fig:robust_vis}.

\section{Conclusion}
\label{Conclusions}
In this paper, we present Mask Guided (MG) Matting, a general framework to resolve the natural image matting problem. Unlike previous methods, our method is not tailored to some specific guidance mask. Instead, it can handle versatile guidance masks such as a trimap, a rough segmentation mask, or a low-quality alpha matte. The key of the robustness of our model lies in the Progressive Refinement Network, which provides self-guidance and progressively refine the uncertain regions during the decoding process. Further, we also propose a simple yet effective method called Random Rendering to resolve the limitation of existing dataset and learn a better foreground color estimation model, which is important yet rarely studied before. Moreover, we release a new real-world matting dataset with high-quality label to better quantitatively evaluate matting models in a real-world scenario, which we hope could shed some light on the direction towards a real-life matting. 

{\small
\bibliographystyle{ieee_fullname}
\bibliography{egbib}
}

\end{document}